\documentclass[letterpaper]{article}
\usepackage{aaai25}
\usepackage{times}
\usepackage{helvet}
\usepackage{courier}
\usepackage[hyphens]{url}
\usepackage{graphicx}
\usepackage{natbib}
\usepackage{caption}
\usepackage{algorithm}
\usepackage{algorithmic}
\usepackage{acronym}
\usepackage[skip=3pt,font=small]{subcaption}
\usepackage{amsmath,amsfonts,mathtools}
\usepackage[dvipsnames,svgnames,x11names]{xcolor}
\usepackage{booktabs,tabularx,colortbl,multirow,array,makecell}
\usepackage[capitalise,nameinlink]{cleveref}
\usepackage{xspace}
\usepackage{cuted}

\usepackage{newfloat}
\usepackage{listings}
\DeclareCaptionStyle{ruled}{labelfont=normalfont,labelsep=colon,strut=off} 
\floatstyle{ruled}
\newfloat{listing}{tb}{lst}{}
\floatname{listing}{Listing}

\definecolor{WeiColor}{rgb}{1,0.33,0.64}
\definecolor{myDarkBlue}{RGB}{55,81,139}
\definecolor{myLightBlue}{RGB}{158,186,217}
\definecolor{myLightGreen}{RGB}{126,171,85}
\definecolor{myRed}{RGB}{176,36,24}
\definecolor{QiColor}{RGB}{81,139,55}
\definecolor{myBlue}{RGB}{64,224,205}
\definecolor{red}{RGB}{255,0,0}
\definecolor{lightyellow}{RGB}{227,207,87}
\definecolor{lightred}{rgb}{0.5, 0.0, 0.13}
\definecolor{lightpurple}{rgb}{0.74, 0.2, 0.64}

\makeatletter
\DeclareRobustCommand\onedot{\futurelet\@let@token\@onedot}
\def\@onedot{\ifx\@let@token.\else.\null\fi}
\def\eg{\emph{e.g}\onedot} 

\def\ie{\emph{i.e}\onedot}

\makeatother

\crefname{algorithm}{Alg.}{Algs.}
\Crefname{algocf}{Alg.}{Algs.}
\crefname{section}{Sec.}{Secs.}
\Crefname{section}{Section}{Sections}
\crefname{table}{Tab.}{Tabs.}
\Crefname{table}{Table}{Tables}
\crefname{figure}{Fig.}{Fig.}
\Crefname{figure}{Figure}{Figure}

\newcommand{\task}{\text{FloNa}\xspace}
\newcommand{\model}{\text{FloDiff}\xspace}
\newcommand\supp{\textit{supplementary material}\xspace}

\acrodef{flona}[\task]{\underline{Flo}or Plan Visual \underline{Na}vigation}
\acrodef{vint}[ViNT]{Visual Navigation Transformer}
\acrodef{nomad}[NoMaD]{Navigation with Goal Masked Diffusion}
\acrodef{mlp}[MLP]{Multi Layer Perceptron}
\acrodef{mse}[MSE]{mean squared error}

\title{FloNa: Floor Plan Guided Embodied Visual Navigation}
\author{
    Jiaxin Li\textsuperscript{\rm 1},
    Weiqi Huang\textsuperscript{\rm 1},
    Zan Wang\textsuperscript{\rm 1},
    Wei Liang\textsuperscript{\rm 1, 2,\textrm{\dag}},
    Huijun Di\textsuperscript{\rm 1,\textrm{\dag}},
    Feng Liu\textsuperscript{\rm 3}
}
\affiliations{
    \textsuperscript{\textrm{\dag}} indicates corresponding authors \quad{}
    \textsuperscript{\rm 1}Beijing Institute of Technology, Beijing, China\\
    \textsuperscript{\rm 2}Yangtze Delta Region Academy of Beijing Institute of Technology, Jiaxing, China\\
    \textsuperscript{\rm 3}Beijing Racobit Electronic Information Technology Co., Ltd.\\
    \{lijiaxin, huangweiqi, wangzan, ajon, liangwei\}@bit.edu.cn, liufeng@racobit.com
}

\begin{document}

\maketitle

\begin{strip}%
    \vspace{-52pt}
    \centering
    \captionsetup{type=figure}
    \includegraphics[width=\textwidth]{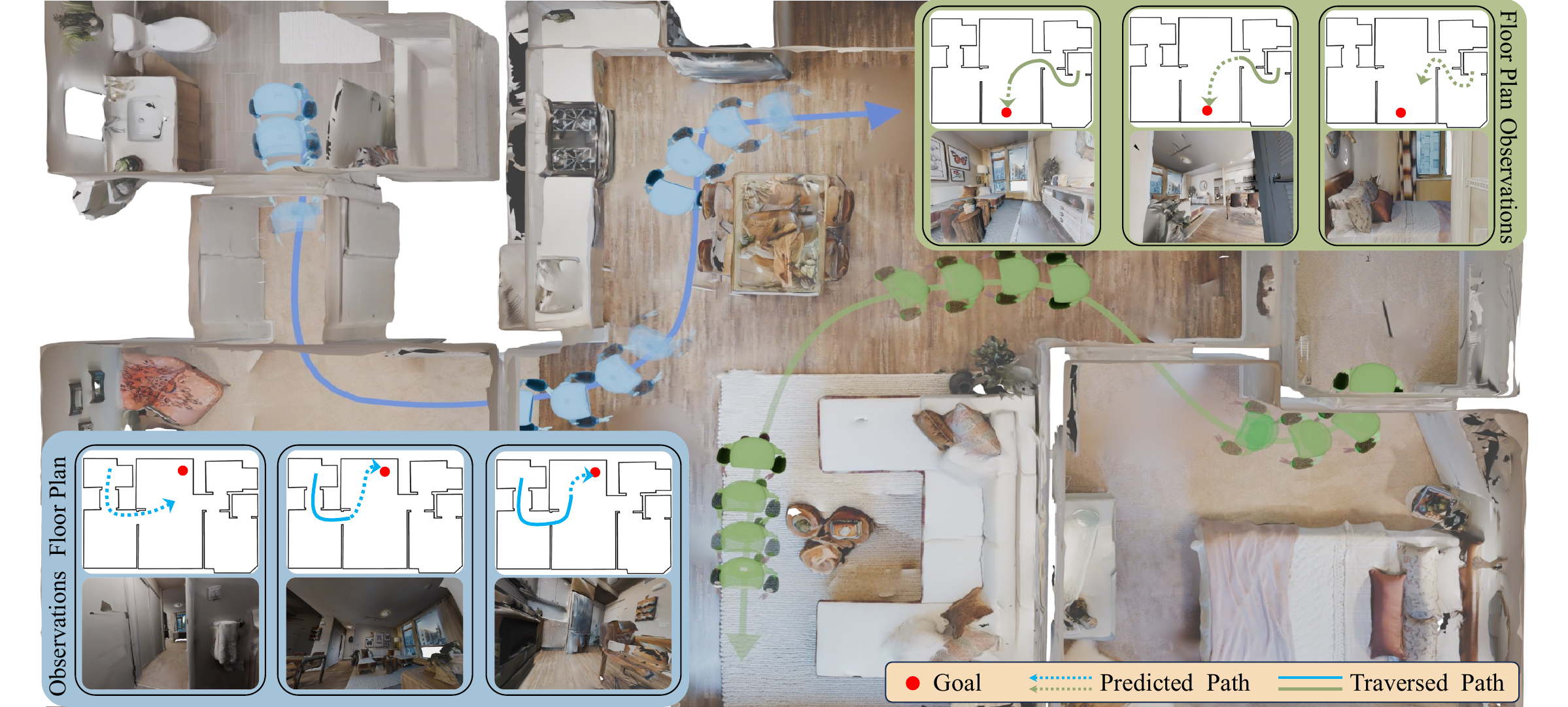}
    \captionof{figure}{\textbf{\ac{flona}:} Given a floor plan with a marked goal indicated by the \textcolor{red}{red} dot, the agent's task is to navigate to the corresponding target location in the environment using RGB observations. To tackle this task, we propose \model, a novel diffusion policy-based framework that iteratively generates and refines the planned trajectory.
    }
    \label{fig:teaser}
\end{strip}

\begin{abstract}
Humans naturally rely on floor plans to navigate in unfamiliar environments, as they are readily available, reliable, and provide rich geometrical guidance. However, existing visual navigation settings overlook this valuable prior knowledge, leading to limited efficiency and accuracy. To eliminate this gap, we introduce a novel navigation task: \textbf{\acf{flona}}, the first attempt to incorporate floor plan into embodied visual navigation.
While the floor plan offers significant advantages, two key challenges emerge: (1) handling the spatial inconsistency between the floor plan and the actual scene layout for collision-free navigation, and (2) aligning observed images with the floor plan sketch despite their distinct modalities.
To address these challenges, we propose \model, a novel diffusion policy framework incorporating a localization module to facilitate alignment between the current observation and the floor plan. We further collect $20k$ navigation episodes across $117$ scenes in the iGibson simulator to support the training and evaluation. Extensive experiments demonstrate the effectiveness and efficiency of our framework in unfamiliar scenes using floor plan knowledge.
Project website: \textcolor{magenta}{\url{https://gauleejx.github.io/flona/}}.
\end{abstract}

\section{Introduction}
An essential task in embodied AI is enabling agents to navigate in diverse environments toward a goal, represented as points \cite{savva2019habitat}, images \cite{zhu2017target,shah2023vint}, objects \cite{chaplot2020object}, or language instructions \cite{anderson2018vision,wang2023active}. Recently, researchers have increasingly leveraged easily accessible prior knowledge \cite{shah2021ving} to improve efficiency and accuracy. Floor plans, in particular, serve as a valuable and widely available source of such knowledge, offering high-level semantic and geometric information that can assist agents in localizing and navigating in unfamiliar spaces. Moreover, incorporating the floor plan benefits various applications such as emergency response, search and rescue, and pathfinding in dynamic public environments.

Previous studies have explored integrating floor plans to facilitate localization and navigation, often relying on multi-sensor fusion \cite{li2021cognitive} or imposing constraints on the floor plan structure \cite{boniardi2019robot}, which limits their practical applicability. Inspired by the human ability to navigate efficiently in unfamiliar environments using minimal abstract information and visual cues, we aim to reduce both sensor dependence and structural constraints on the floor plan in navigation.

In this work, we introduce a novel navigation task, \textbf{\acf{flona}}, where agents navigate within an environment using an abstract floor plan and a series of RGB observations, as depicted in \cref{fig:teaser}. Despite the valuable prior knowledge provided by the floor plan, this task remains challenging for two primary reasons. First, spatial inconsistency exists due to the significant difference between the floor plan and the actual observed layout, primarily caused by furniture placement in the scenes. This inconsistency can lead to collisions during navigation. Second, observation misalignment arises because the floor plan offers abstract topological information, whereas the RGB observations capture the appearance of natural scenes from a specific viewpoint. Such misalignment results in incorrect localization of the current observation within the floor plan, thereby hindering the effectiveness of planning.

To address these challenges, we develop \model, a novel diffusion policy framework that exploits strong action distribution modeling capability to learn to handle the spatial inconsistency from extensive demonstrations implicitly. \model also integrates an explicit localization module to align the observation and floor plan, resulting in two variants based on how current agent pose is derived: (1) Naive-\model, which learns to predict the pose during training, and (2) Loc-\model, which directly uses either ground truth poses or predictions from a pre-trained model. Both variants employ a Transformer backbone to fuse the floor plan and current observation, followed by passing the fusion into a policy network that learns to denoise the action sequence.

For benchmarking, we collect a dataset comprising approximately $20k$ navigation episodes across $117$ distinct scenes using the iGibson simulator \cite{li2021igibson}. The dataset includes around $3.3M$ images captured with a $45$-degree field of view. We split the scenes into $67$ for training and $50$ for testing to assess the model's generalization capability to unseen environments. Each scene comprises a floor plan, a traversability map, and sufficient navigation episodes. Each episode contains an A*-generated trajectory paired with corresponding RGB observations.

Extensive experiments demonstrate the effectiveness and efficiency of our method in navigating within unseen environments using a floor plan. Compared to baseline methods, our method achieves a higher Success Rate (SR) and greater efficiency, as measured by Success Weighted by Path Length (SPL). Additionally, we comprehensively analyze \model's different capabilities, including localization, collision avoidance, planning with diverse goals, and robustness. Real-world deployment on an Automated Guided Vehicle (AGV) without finetuning further highlights its robustness and generalization, proving its potential to handle unseen scenarios effectively in practical settings.

Our contributions are summarized as follows:
\begin{itemize}
    \item We introduce \task, a novel task of navigating toward a goal using RGB observations and a floor plan, enriching the application scenarios in embodied visual navigation.
    \item We propose a novel end-to-end diffusion policy-based framework, \ie, \model, incorporating explicit agent localization to solve \task efficiently and effectively.
    \item We conduct extensive experiments on our curated dataset and thoroughly analyze \model's capabilities across various dimensions, demonstrating its superiority over baseline methods.
\end{itemize}

\section{Related Work}
\subsection{Visual Navigation}
Visual navigation aims at enabling robots or autonomous agents to navigate through environments using visual information. The definition of navigation tasks has evolved alongside the advancements in the field. \citet{anderson2018evaluation} are the first to explicitly categorize three navigation tasks based on goal types: PointGoal, ObjectGoal, and AreaGoal. The PointGoal task requires the agent to navigate to a specific location, emphasizing the self-localization capabilities \cite{wijmans2019dd,ramakrishnan2021habitat,zhao2021surprising,partsey2022mapping,tang2022monocular}. In contrast, the ObjectGoal and AreaGoal tasks involve commonsense knowledge of the environment, such as recognizing objects and understanding where they are typically located \cite{yadav2023ovrl,ramakrishnan2022poni,gadre2023cows}. With advancements in image and natural language understanding, the field has shifted focus toward navigation tasks guided by high-level semantics. Tasks like image navigation \cite{zhu2017target,krantz2023navigating,yadav2023ovrl,chaplot2020neural} and language navigation \cite{anderson2018vision,song2023llm, min2021film,wang2023dreamwalker,wang2023active} have emerged, filling gaps in various real-world and natural application scenarios. Beyond vision and language, navigation using other modalities, such as sound, has also been explored \cite{chen2020soundspaces,gan2020look,chen2021learning}.
Despite the promising progress in prior studies, efficient and effective exploration remains a critical challenge in tasks involving unknown environments. In this work, we explore the potential of leveraging readily available prior information, \ie, floor plans, to mitigate the dependence on time-consuming exploration, further enhancing the effectiveness and efficiency of navigation.

\subsection{Floor Plan for Localization and Navigation}
Floor plans, as stable structural information, have been widely explored in localization and navigation tasks. One line of research approaches localization within a traditional optimization framework, leveraging floor plans in combination with various sensor signals such as LiDAR \cite{boniardi2019pose,boniardi2017robust,li2020online,mendez2020sedar,wang2019glfp}, images \cite{ito2014w,boniardi2019robot}, or visual odometry \cite{chu2015you}. Another branch adopts learning-based methods \cite{howard2021lalaloc,howard2022lalaloc++,min2022laser,chen2024f3loc} to solve the localization task. Some studies have also attempted to use augmented topological maps generated from sketch floor plans \cite{setalaphruk2003robot} or architectural floor plans \cite{li2021cognitive} to aid the navigation. These works highlight the benefits of incorporating prior knowledge from floor plans for indoor localization and navigation. However, their dependence on multiple sensors hinders the practical applicability of the approach. To overcome this limitation, we investigate the feasibility of accomplishing the navigation task using only RGB observations and floor plans in this work.

\section{Task Setting}
\begin{figure*}[t!]
    \centering
    \includegraphics[width=\textwidth]{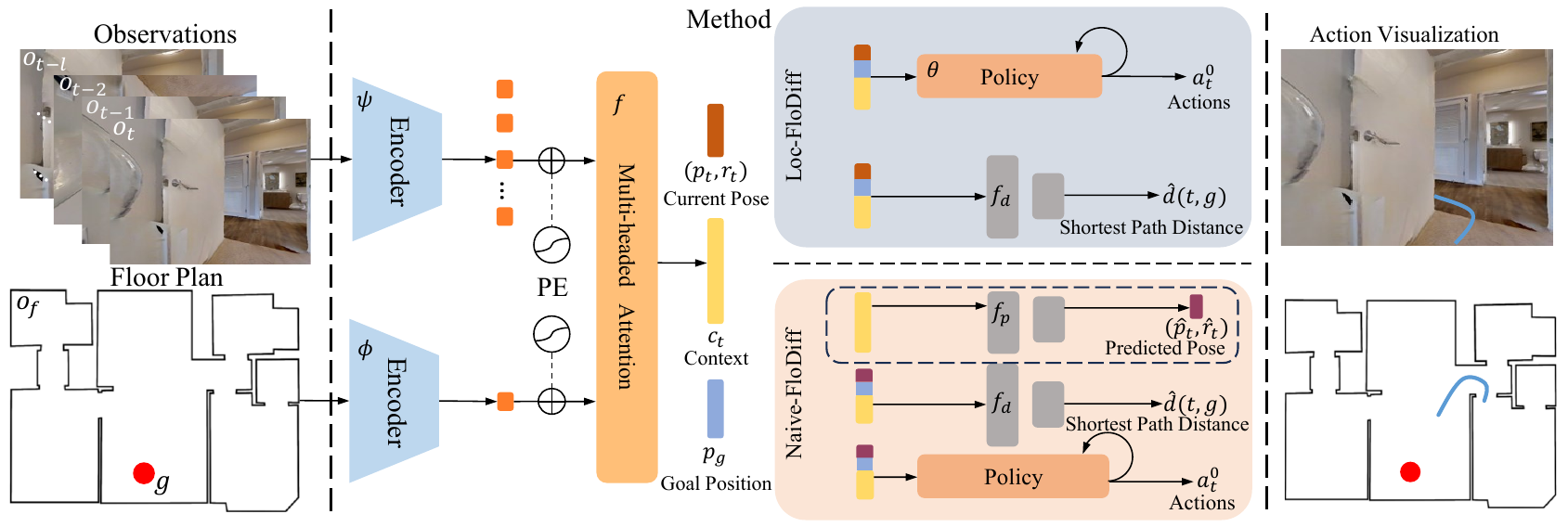}
    \caption{\textbf{Pipeline overview.} \model employs an attention module to fuse features from visual observation and floor plan, yielding a context embedding $c_t$. Depending on how the current agent pose is derived, \model has two variants: (1) Naive-\model (below), which learns to predict the current pose $(\hat{p}_t, \hat{r}_t)$ during policy learning; (2) Loc-\model (above), which directly uses the ground truth pose or predictions from pre-trained models. The concatenation of the observation context $c_t$, goal position $p_g$, and current agent pose $(p_t, r_t)$ is then fed into the policy network to generate actions.}
    \label{fig:pipeline}
    \vspace{-5pt}
\end{figure*}

\paragraph{Task Definition} The goal of \task requires embodied agents to navigate from a starting position to a specific goal position in a 3D environment. During the navigation, the agent can only receive the ego-centric RGB image, the environment's floor plan image, and a goal position, represented as a red dot in \cref{fig:pipeline}.
Specifically, given the floor plan $o_f$, the agent receives the observation image $o_t$ at each timestep $t$ and produces action $a_t$ to reach the goal $g$. A resulting episode $\{x_0, a_1, x_1, \cdots, a_T, x_T\}$, where $x_i$ comprises the agent's location $p_i$ and orientation $r_i$, is considered successful if the last position $p_T$ is within a specified distance threshold $\tau_d$ from the goal and the number of collisions with the scene does not exceed a given threshold $\tau_c$. $x_0 = \{p_0, r_0\}$ is the starting pose and $T$ is the total step.

\paragraph{Simulator Setup} Following \citet{li2021igibson}, we build the simulation environment upon the iGibson simulator and employ a Locobot as the embodied agent. The agent has a height of $0.85$ meters and a base radius of $0.18$ meters, and is equipped with an RGB camera with a resolution of $512\times512$ pixels. We define the action as $a_t \coloneqq p_{t} - p_{t-1}$, $a_t \in \mathbb{R}^2$, representing continuous movement in the 2D plane.

\begin{figure}[t!]
    \centering
    \includegraphics[width=\linewidth]{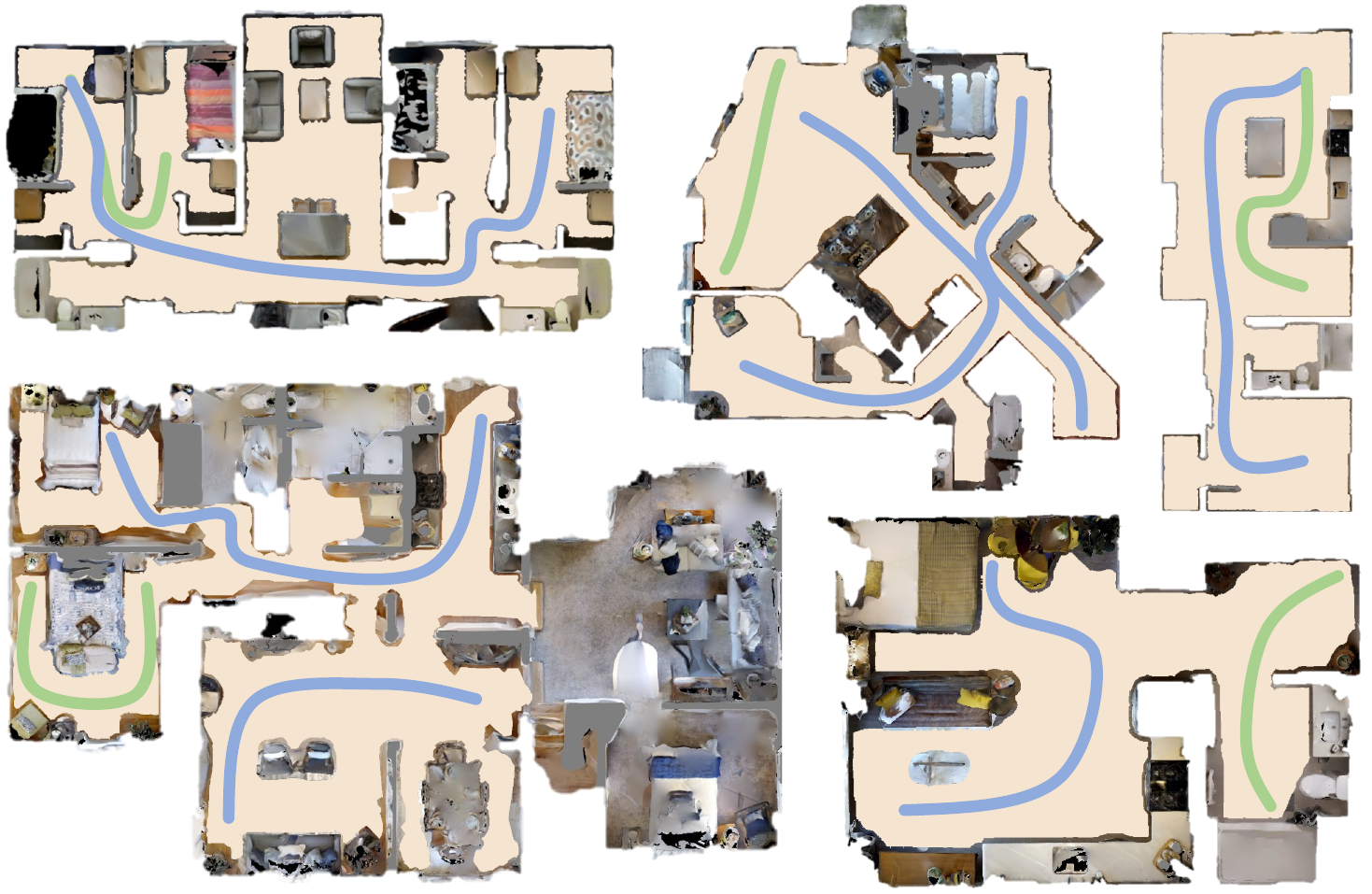}
    \caption{\textbf{Typical scenes, navigable areas, navigation episodes in our collected dataset}. We show the short and long episodes using \textcolor{myLightGreen}{green} and \textcolor{myLightBlue}{blue} colors, respectively.}
    \label{fig:dataset_examples}
\end{figure}

\section{Dataset Design}
To facilitate the benchmarking, we collect a large-scale dataset comprising $20,214$ navigation episodes across $117$ static indoor scenes from Gibson \cite{xia2018gibson}, resulting in a total of $3,312,480$ images captured with a $45$-degree field of view. The Gibson scenes are reconstructed from the homes and offices with a Matterport device, which preserves the textures observed by the sensor, thus minimizing the sim-to-real gap. We split the dataset into training and testing sets, which comprise $67$ scenes and $50$ scenes, respectively.

For each scene, we provide a floor plan and a navigable map. We directly adopt the manually-annotated floor plans from \citet{chen2024f3loc}. For navigable maps, we employ a coarse-to-fine approach to obtain them by executing the following two steps. First, we generate a rough one from the scene's mesh. Then, we manually refine it to address scanning artifacts; for example, the poorly reconstructed chair legs may lead to a navigable area. Based on scene size, we collect $150$ episodes in small scenes, $180$ in medium scenes, and $200$ in large scenes. Before collecting each episode, we randomly sample two points at least $3$ meters apart as the starting and target points. We then use the A* algorithm to search for a collision-free trajectory from the start to the target. We record the trajectory positions in the pre-defined world coordinate system and render the observation images during the search process. The episode trajectory length ranges from $4.53$ to $42.03$ meters. \cref{fig:dataset_examples} presents some trajectory examples. Please refer to \supp for more details about the dataset construction.

\section{Preliminary}
\subsection{Diffusion Model}
The diffusion model \cite{ho2020denoising, sohl2015deep} is a probabilistic framework used in generative tasks. It learns to gradually denoise the sampled Gaussian noise to generate desired data through an iterative denoising process. Specifically, beginning with sampled noise $x^K$, the model performs $K$ iterations of denoising to produce a series of intermediate poses $x^{K-1}, \cdots, x^0$. The process can be denoted as:
\begin{equation}
    x^{k-1} = \alpha(x^k - \gamma\epsilon_\theta(x^k,k) + \sigma z),
    \label{eq:ddpm}
\end{equation}
where $\alpha$, $\gamma$, $\sigma$ are noise schedule functions, $\epsilon_\theta$ is the noise prediction network with parameters $\theta$ and $z \sim \mathcal{N}(0, \mathbf{I})$.

\subsection{Diffusion Policy}
Diffusion policy \cite{chi2023diffusion} extends the concept of diffusion models, which are typically used for content generation, to policy learning, allowing an agent to make decisions based on learned probabilistic distributions. The policy usually incorporates an encoder to transform the observation into a lower-dimensional feature, serving as the condition for the denoising process. \cref{eq:ddpm} can be adapted to:
\begin{equation}
    \mathcal{A}^{k-1}_t = \alpha(\mathcal{A}^{k}_t - \gamma\varepsilon_\theta(\mathcal{A}^{k}_t,\mathcal{O}_t,k)+ \sigma z),
    \label{eq:dp}
\end{equation}
where $\mathcal{A}_t$ and $\mathcal{O}_t$ denote the action and the observation at each time step $t$, respectively.

\section{Method}
Building upon diffusion policy, we propose a novel end-to-end framework, \ie, \model, for solving \task. The overall architecture is illustrated in \cref{fig:pipeline}. Next, we will detail the \model framework and the training process.

\subsection{Diffusion Policy for \task}
\paragraph{Framework Overview}
\model consists of a transformer-based backbone that fuses visual observations with the floor plan and a policy network designed to navigate within the scene. As discussed, an ideal agent must be capable of (1) navigating to the goal without colliding with the scene and (2) aligning visual observations with the floor plan. To address the first challenge, we employ the powerful diffusion policy to learn navigation in crowded scenes from extensive demonstrations. For the second challenge, we explicitly predict the agent pose from observations or utilize a pre-trained model for pose prediction.

\paragraph{Observation Context}
In our work, the visual observation $\mathcal{O}_{t}$ contains the observed images $\{o_{t-l}, o_{t-l+1}, ..., o_t\}$, where $l$ denotes the context length, and the floor plan image $o_f$. Inspired by a wealth of prior work on training high-capacity policies \cite{sridhar2023nomad,shah2023vint}, we employ a transformer-based backbone to process the observation context. Specifically, for each observed image $o_i$, $i=t-l, \cdots, t$, we employ a shared EfficientNet-B0 \cite{tan2019efficientnet} as the encoder, denoted as ${\psi}$, to produce the visual latent features independently. Following \ac{vint} \cite{shah2023vint} that encodes the relative difference between the current observation and the goal image, we employ another EfficientNet-B0 as encoder $\phi$ to process the floor plan image. We train both ${\psi}$ and $\phi$ from scratch. Subsequently, we apply multi-headed attention \cite{vaswani2017attention} layers $f(\cdot)$ to fuse the two branch features. The final output feature serves as the observation context vector $c_t$.

\paragraph{Goal Position}
We compute goal position $p_g$ in the world coordinate according to the marked goal point $g$. Precisely, we first determine the pixel coordinates of the goal on the floor plan image, denoted as $u_g$. We then convert the coordinates into the world coordinates using the floor plan resolution $\mu$ defined as the distance per pixel and the offset $\delta$ provided by \citet{chen2024f3loc} through $p_g=\frac{u_g}{\mu} + \delta$.

\paragraph{Policy} 
Based on different ways of obtaining the current agent pose, we propose two variants of \model.

First, as described above, $f(\cdot)$ fuses the observed images with the floor plan, indicating that $c_t$ encodes valuable information for predicting the agent poses. To achieve this, we introduce a fully connected network $f_p$ to predict the agent pose $\hat{x}_t = (\hat{p}_t, \hat{r}_{t})$ from the observation context vector $c_t$. Following \citet{shah2023vint}, we also use an additional fully connected network $f_{d}$ to predict the shortest path distance $\hat{d}(t, g)$ between the current position and the goal, as illustrated in \cref{fig:pipeline}. This design helps mitigate the policy's tendency to move directly toward the goal, which could otherwise lead to dead ends. Finally, we concatenate the observation context vector $c_t$, goal position $p_g$, and predicted agent pose $x_t$ as a conditional vector, from which the diffusion policy $\theta$ learns to denoise the sampled noise and generate the action sequence. We denote this variant as Naive-\model.

Second, we can directly use the ground truth agent pose or predicted pose from a pre-trained model, \ie, F$^3$Loc \cite{chen2024f3loc}. Unlike Naive-\model, this design eliminates the need for $f_p$. We denote such \model variants as Loc-\model, including Loc-\model (GT) and Loc-\model ($\text{F}^3$).

For all \model variants, we model the predicted action sequence $\mathcal{A}_{t}^{0}$ at time step $t$ as the future action sequence with a horizon of $H_p$, where the agent executes the first $H_a$ steps. During inference, we sample the predicted action sequence as described in \cref{eq:dp} and compute the next position by adding the action vector to the current position.

\subsection{Training}
We train \model on the training set, which consists of 67 indoor scenes, encompassing $11,575$ episodes and approximately $26$ hours of trajectory data. During training, we randomly sample trajectory segments with a fixed horizon from the episodes as training samples. To increase the diversity of target goals, we also randomly select a target position behind the sampled trajectory rather than using the episode endpoint as the target. We train Naive-\model in an end-to-end manner using the following loss functions:
\begin{equation}
\begin{aligned}
    \mathcal{L}(\phi, \psi, f, \theta, f_d, &\ f_p) = \texttt{MSE}(\epsilon^k, \epsilon_\theta(c_t, \mathcal{A}_t^0+\epsilon^k,k)) \\
                                            & + \lambda_1\cdot\texttt{MSE}(d(t, g), f_d(c_t, \hat{x}_t, p_g)) \\
                                            & + \lambda_2\cdot\texttt{MSE}(x_t, f_p(c_t)).
\end{aligned}
\end{equation}
Similarly, we train Loc-\model using:
\begin{equation}
\begin{aligned}
    \mathcal{L}(\phi, \psi, f, \theta, &\ f_d) = \texttt{MSE}(\epsilon^k, \epsilon_\theta(c_t, \mathcal{A}_t^0+\epsilon^k,k)) \\
                                       & + \lambda_3\cdot\texttt{MSE}(d(t, g), f_d(c_t, \hat{x}_t, p_g)).
\end{aligned}
\end{equation}
$\lambda_1$, $\lambda_2$, and $\lambda_3$ are hyperparameters for weighting different loss terms. $\texttt{MSE}(\cdot)$ computes mean square error.

In the implementation, \model is trained for $5$ epochs using AdamW \cite{loshchilov2017fixing} optimizer with a fixed learning rate of $0.0001$. We empirically set $\lambda_1 = \lambda_3 = 0.001$ and $\lambda_2 = 0.005$. The attention layers are built using the native PyTorch implementation. The number of multi-head attention layers and heads are both $4$. We set the dimension of the observation context vector $c_t$ to $256$. The diffusion policy is trained using the Square Cosine Noise Scheduler \cite{nichol2021improved} with $K = 10$ denoising steps. The noise prediction network $\epsilon_\theta$ adopts a conditional U-Net architecture following \cite{janner2022diffuser} with $15$ convolutional layers. We set the diffusion horizon as $H_p = 32$ and employ the first $H_a=16$ steps to execute in each iteration. We train \model using one NVIDIA RTX3090 GPU and assign a batch size of $256$.

\section{Experiments}
This section begins with an introduction to the experimental setup, including the baseline methods and the evaluation metrics. We then present the results and the performance analysis regarding localization, collision avoidance, planning, and robustness, followed by a real-world deployment.

\subsection{Experimental Setting}
We evaluate our proposed method on $50$ test indoor scenes. $10$ pairs of start and end points are randomly selected in each scene, totaling $500$ test pairs. \cref{tab:test_set_stat} presents statistics on the straight-line distance and travel distance between the start and end points of the $500$ test episodes. Collisions may occur because the floor plan does not account for furniture obstacles. To mitigate this, we adopt a strategy in which the agent will rotate $45$ degrees clockwise and re-predict future actions when a collision happens.

\begin{table}[t!]
    \centering
    \caption{\textbf{Statistics of test episodes.} The units are in meters.}
    \vspace{-2pt}
    \label{tab:test_set_stat}
    \begin{tabular}{ccccc}%
        \toprule
        & Min & Max & Mean & Median \\
        \midrule
        straight-line distance & 5.89 & 29.87 & 7.47 & 6.51 \\
        travel distance        & 6.97 & 34.72 & 9.15 & 8.11 \\
        \bottomrule
    \end{tabular}%
\end{table}

\begin{table*}[t!]
    \centering
    \caption{\textbf{SR ($\%$) and SPL ($\%$) of different models under various conditions}. The \textbf{bold} number indicates the best result, and the \underline{underlined} number represents the second-best result.}
    \vspace{-2pt}
    \label{tab:results}
    \resizebox{\linewidth}{!}{%
    \begin{tabular}{cccccccccccccccccc}%
        \toprule
        \multicolumn{3}{c}{\multirow{3}[4]{*}{Method}}& \multicolumn{3}{c}{Loc-A* (F$^3$)}            & \multicolumn{3}{c}{Loc-A* (GT)} &  \multicolumn{3}{c}{Naive-\model}&
        \multicolumn{3}{c}{Loc-\model (F$^3$)} & \multicolumn{3}{c}{Loc-\model (GT)}  \\
        \cmidrule{4-18}
         & &                                          & \multicolumn{3}{c}{$\tau_d (m)$}                 & \multicolumn{3}{c}{$\tau_d (m)$}    & \multicolumn{3}{c}{$\tau_d (m)$} & \multicolumn{3}{c}{$\tau_d (m)$} & \multicolumn{3}{c}{$\tau_d (m)$} \\
        \cmidrule(rr){4-6}\cmidrule(lr){7-9}\cmidrule(lr){10-12}\cmidrule(lr){13-15}\cmidrule(l){16-18}
         & &                                          & 0.25 & 0.30 & 0.35                               & 0.25 & 0.30 & 0.35                 & 0.25 & 0.30 & 0.35                 & 0.25 & 0.30 & 0.35                & 0.25 & 0.30 & 0.35              \\
        \midrule
            \multicolumn{1}{c|}{\multirow{4}{*}{SR(\%) $\uparrow$}}  &\multicolumn{1}{c}{\multirow{4}{*}{$\tau_c$}}     & 10       & 0.00 & 0.60 & 0.60    & \underline{16.6} &\underline{25.4} & \underline{38.6}        & 4.20 & 5.00 & 5.60  & 4.80 & 6.00 & 6.40    & \textbf{39.0} & \textbf{44.2} & \textbf{44.6}\\
        \multicolumn{1}{c|}{}                         &\multicolumn{1}{c}{}                              & 30       & 0.00 & 0.60 & 0.60    & \underline{18.2} & \underline{28.4} & \underline{43.8}        & 9.20 & 11.8 & 13.0 & 11.6 & 14.0 & 15.0    & \textbf{53.4} & \textbf{60.8} & \textbf{61.0}\\
        \multicolumn{1}{c|}{}                         &\multicolumn{1}{c}{}                              & 50       & 0.00 & 0.60 & 0.60    & \underline{18.5} & \underline{29.4} & \underline{45.6}        & 12.0 & 15.6 & 17.2 & 17.8 & 20.6 & 22.4    & \textbf{59.2} & \textbf{68.4} & \textbf{68.4}\\
        \multicolumn{1}{c|}{}                         &\multicolumn{1}{c}{}                              & $\infty$ & 0.00 & 0.60 & 0.60    & 18.6 & 29.6 & \underline{45.8}        & 20.2 & 24.8 & 27.0   & \underline{26.8} & \underline{30.8} & 33.0    & \textbf{66.0} & \textbf{75.8} & \textbf{76.0}\\             
        \midrule
        \multicolumn{1}{c|}{\multirow{4}{*}{SPL(\%) $\uparrow$}} &\multicolumn{1}{c}{\multirow{4}{*}{$\tau_c$}}     & 10       & 0.00 & 0.59 & 0.59    & \underline{16.5} & \underline{25.3} & \underline{38.51}       & 3.31 & 4.15 & 4.68  & 4.02 & 5.24 & 5.54    & \textbf{36.3} & \textbf{41.4} & \textbf{42.1}\\
        \multicolumn{1}{c|}{}                         &\multicolumn{1}{c}{}                              & 30       & 0.00 & 0.59 & 0.59    & \underline{18.1} & \underline{28.3} & \underline{43.51}       & 4.92 & 6.46 & 7.26    & 6.40 & 8.25 & 8.98    & \textbf{44.1} & \textbf{50.4} & \textbf{51.2}\\
        \multicolumn{1}{c|}{}                         &\multicolumn{1}{c}{}                              & 50       & 0.00 & 0.60 & 0.60    & \underline{18.5} & \underline{29.3} & \underline{45.11}     & 5.54 & 7.31 & 8.30    &8.07 & 9.98 & 10.8    & \textbf{46.6} & \textbf{53.4} & \textbf{54.1} \\
        \multicolumn{1}{c|}{}                         &\multicolumn{1}{c}{}                              & $\infty$ & 0.00 & 0.60 & 0.60    & \underline{18.5} & \underline{29.4} & \underline{45.31}      & 6.82 & 8.71 & 9.74  & 9.58 & 11.7 & 12.5    & \textbf{48.2} & \textbf{55.2} & \textbf{56.0} \\
        \bottomrule
    \end{tabular}%
    }%
\end{table*}

\paragraph{Baselines}
In our evaluation, in addition to Naive-\model and Loc-\model, we further design a modular baseline including a localization module and an explicit path-planning module, denoted as Loc-A*. The localization modules in both Loc-\model and Loc-A* are implemented in two distinct ways, \ie, using GT or F$^3$Loc \cite{chen2024f3loc}, comparing their effectiveness across different settings. The five methods are described as follows:
\begin{itemize}
    \item \textbf{Loc-A* (F$^3$):} In this baseline, the agent localizes itself using pre-trained F$^3$Loc \cite{chen2024f3loc}. Subsequently, the A* algorithm is employed to plan a trajectory from the predicted position to the goal on the floor plan and execute the mapped actions in the environment.
    \item \textbf{Loc-A* (GT):} Unlike \textbf{ Loc-A* (F$^3$)}, this baseline replaces the predicted pose with the ground truth pose.
    \item \textbf{Naive-\model:} This method directly learns to predict the agent pose during policy training.
    \item \textbf{Loc-\model (F$^3$):} This method is based on our Loc-\model model and utilizes the same localization module as \textbf{Loc-A* (F$^3$)}.
    \item \textbf{Loc-\model (GT):} Unlike \textbf{ Loc-\model (F$^3$)}, the predicted poses are replaced by the ground truth poses.
\end{itemize}

\paragraph{Metrics}
The performance of the above navigation methods is evaluated using two primary metrics, \ie,

\begin{itemize}
    \item Success Rate (SR): ${\rm SR} = \frac{1}{N} \sum_{i = 1}^{N}S_i$, where $N$ is the total number of test episodes. $S_i$ equals $1$ if episode $i$ is successful; otherwise, it equals $0$.
    \item Success Weight by Path Length (SPL): ${\rm SPL} = \frac{1}{N} \sum_{i = 1}^{N}S_i \cdot \frac{l_i}{\max(p_i,l_i)}$, where $p_i$ is the length of the agent's path, and $l_i$ is the shortest path. The metric measures the efficiency of navigation.
\end{itemize}

Recall that a navigation task is successful if the final position is within a specified distance threshold $\tau_d$ from the goal and the number of collisions does not exceed a given threshold $\tau_c$. Since the maximum shortest path length in the test set is $34.72$ meters, any trajectory exceeding $100$ is deemed a failure during the evaluation.

\subsection{Results and Discussions}
\paragraph{Main Results}  
We present the main comparison results in \cref{tab:results}, computing the SR and SPL under varying $\tau_c$ and $\tau_d$. We observe that Loc-\model (GT) consistently outperforms all other methods across varying conditions in both SR and SPL, indicating that the proposed diffusion policy-based method can effectively solve the \ac{flona}. Below, we analyze the methods' performance across various dimensions, including localization, collision avoidance, planning, and robustness. We provide additional experimental results and qualitative visualization in \supp.

\begin{table}[t!]
    \centering
    \caption{\textbf{Mean collision counts.} A small number is better.}
    \vspace{-2pt}
    \label{tab:collison_nums}
    \begin{tabular}{cccc}
        \toprule
        Method             & $\tau_c=10$   & $\tau_c=30$    & $\tau_c=50$     \\
        \midrule
        Loc-A* (F$^3$)     & 9.94          & 29.82          & 49.71           \\
        Loc-A* (GT)        & \textbf{6.03} & 16.21          & 25.65           \\
        Naive-\model       & 9.66          & 28.02          & 45.26           \\
        Loc-\model (F$^3$) & 9.52          & 27.40          & 43.66           \\
        Loc-\model (GT)    & 6.44          & \textbf{15.57} & \textbf{22.45}  \\
        \bottomrule
    \end{tabular}%
\end{table}

\paragraph{Localization Analysis} 
The results in \cref{tab:results} show that the success rate of Naive-\model is lower than both Loc-\model (GT) and Loc-\model (F$^3$). This suggests that the modular method of decoupling the localization module is more effective in tackling \ac{flona} compared to the end-to-end method. It appears that the encoders face challenges in simultaneously encoding information for both localization and planning.

Notably, Loc-A* (F$^3$) and Loc-\model (F$^3$) perform worse than their ground truth counterparts. Upon reviewing the planned trajectories, we observe that agents using F$^3$Loc often become stuck due to collisions with obstacles in confined areas. This issue is primarily caused by inaccurate localization, which results in unreasonable path planning. We suppose this unsatisfactory localization stems from using F$^3$Loc, which is pre-trained on data collected from an agent with a height of $1.70m$,  significantly different from our agent with $0.85m$ height. We also find that, even under similar unsatisfactory localization conditions, the success rate of Loc\model (F$^3$) does not degrade as severely as that of Loc-A*. Furthermore, Loc-\model (F$^3$) even performs better than Loc-A* (GT) when $\tau_c=\infty$ and $\tau_d=0.25m$ or $\tau_d=0.3m$. This result suggests that while Loc-\model is affected by localization accuracy, it demonstrates a certain robustness to noisy input, allowing it to perform reasonably even under less accurate localization.

\begin{figure}[t!]
    \begin{subfigure}{0.49\linewidth}
        \includegraphics[width=\linewidth]{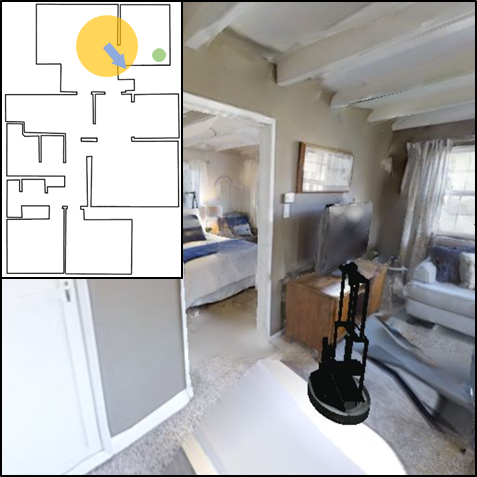}
        \caption{Noisy pose}
        \label{fig:noisy_0}
    \end{subfigure}%
    \hspace{0.01\linewidth}
    \begin{subfigure}{0.49\linewidth}
        \includegraphics[width=\linewidth]{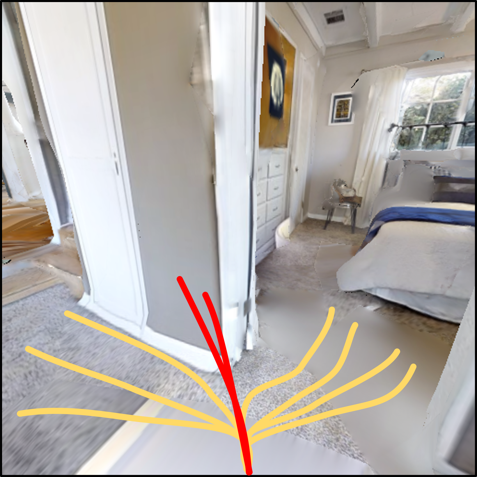}
        \caption{Predicted actions}
        \label{fig:noisy_1}
    \end{subfigure}%
    \caption{\textbf{Robustness of Loc-\model (GT).} (a) The blue arrow is the ground truth agent pose, and the yellow circle indicates the noisy poses. (b) Our method can generate diverse collision-free paths (in \textcolor{Dandelion}{yellow}), even given the noisy poses. The \textcolor{red}{red} collision paths are generated by Loc-A* (GT).}
    \label{fig:noisy}
\end{figure}

\paragraph{Collision Avoidance}
To quantitatively evaluate the collision avoidance performance, we present the mean collision counts of successful cases with $\tau_d = 0.3m$ under varying $\tau_c$ in \cref{tab:collison_nums}. The results demonstrate that Loc-\model (GT) outperforms Loc-A* (GT), and Loc-\model (F$^3$) also results in fewer collisions than Loc-A* (F$^3$). These findings suggest that, when using the same localization module, Loc-\model collides less than Loc-A*, highlighting its collision avoidance capability during navigation. 

To qualitatively assess the collision avoidance capability of Loc-\model, we conduct a diagnostic experiment using noisy ground truth localization. Specifically, the agent is initialized in front of a door, facing the door frame, as shown in \cref{fig:noisy_0}. The agent's goal is to navigate to the nearest room through the door, with the green goal marked on the floor plan. The position $p_t$ input to Loc-\model is sampled from Gaussian noise centered on the ground truth position, with a variance of $1m$, represented by the yellow circle in \cref{fig:noisy_0}. The orientation input $r_t$ is sampled from a uniform distribution over $360^\circ$. Despite the noisy input, the agent successfully predicts collision-free paths most of the time, as shown with the yellow lines in \cref{fig:noisy_1} (although some predicted actions caused the agent to move away from the goal). In contrast, Loc-A* frequently predicts trajectories with collision, as indicated by the red lines in \cref{fig:noisy_1}.

\begin{figure}[t!]
    \centering
    \begin{subfigure}{0.49\linewidth}
        \includegraphics[width=\linewidth]{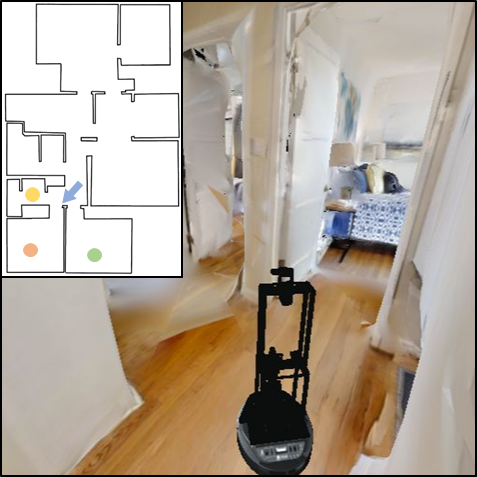}
        \caption{Floor plan with $3$ goals}
        \label{fig:plan_0}
    \end{subfigure}%
    \hspace{0.01\linewidth}
    \begin{subfigure}{0.49\linewidth}
        \includegraphics[width=\linewidth]{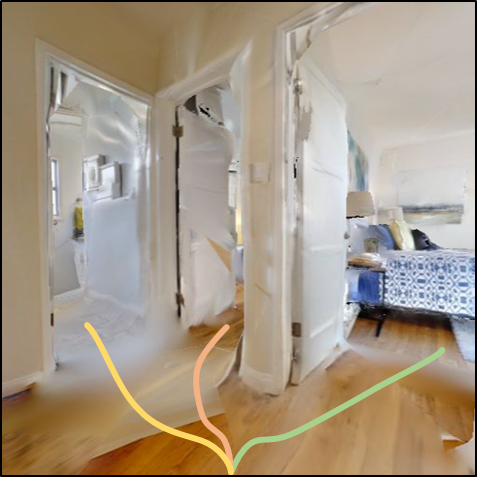}
        \caption{Predicted actions}
        \label{fig:plan_1}
    \end{subfigure}%
    \caption{\textbf{Agent behavior varies given different goals.} (a) The agent starts from the same position but with three different goals. (b) Our approach predicts three distinct paths, each corresponding to a specific goal, as indicated by the respective colors.}
    \label{fig:plan}
\end{figure}

\paragraph{Planning}
The SR and SPL results in \cref{tab:results} demonstrate that Loc-\model (GT) achieves superior performance, highlighting its effectiveness and efficiency in path planning. To qualitatively assess the planning ability of Loc-\model (GT), we position the agent in front of three rooms, as shown in \cref{fig:plan_0},  with the blue arrow indicating its position and heading direction. We assign three goal positions to the agent, each in a separate room and marked by a distinct color on the floor plan in \cref{fig:plan_0}. As shown in \cref{fig:plan_1}, the resulting action sequences, which align with the goal colors, successfully guide the agent to the correct rooms.

\begin{table}[t!]
    \centering
    \caption{\textbf{Results of Loc-\model (GT) with noisy ground truth pose under different noise level.}}
    \vspace{-2pt}
    \label{tab:robustness}
        \begin{tabular}{ccccc}
            \toprule
            Noise Variance        &$0.1m$&$0.3m$&$0.5m$& GT   \\
            \midrule
            SR ($\%$) $\uparrow$  & 26.0 & 20.0 & 24.0 & 68.4 \\
            SPL ($\%$) $\uparrow$ & 10.8 & 8.21 & 12.2 & 53.4 \\
            \bottomrule
        \end{tabular}%
\end{table}

\paragraph{Robustness}
As previously mentioned, \cref{fig:noisy} qualitatively demonstrates that our method can robustly predict collision-free actions most of the time, even with noisy position inputs. To further evaluate the robustness of our method quantitatively, we conduct three experiments in which the Loc-\model (GT) agent navigates to the goal with noisy pose inputs. In each experiment, the agent receives poses with Gaussian noise at a specific variance, \ie, $var = 0.1m$, $var=0.3m$, and $var=0.5m$. All three experiments are evaluated with $50$ cases, with one case per scene from the test set. The results, including SR and SPL metrics under the conditions $\tau_c=50$ and $\tau_d=0.3m$, are presented in \cref{tab:robustness}. Although adding noise to the input pose leads to a decline in both SR and SPL, the performance does not degrade continuously as the noise variance increases. This stability highlights the model's capability of handling uncertainties in localization while maintaining navigation performance.

\subsection{Real-world Deployment}
To validate the effectiveness of the model, we deploy our policy on an AGV. 
We observe that the agent surprisingly completes the navigation tasks in an \textbf{unseen real-world} environment even without finetuning, demonstrating its robustness and generalization capabilities. \textbf{We recommend referring to our project website for the demonstration video of the planning results.}

\begin{figure}[t!]
    \centering
    \begin{subfigure}{0.36\linewidth}
        \includegraphics[width=\linewidth]{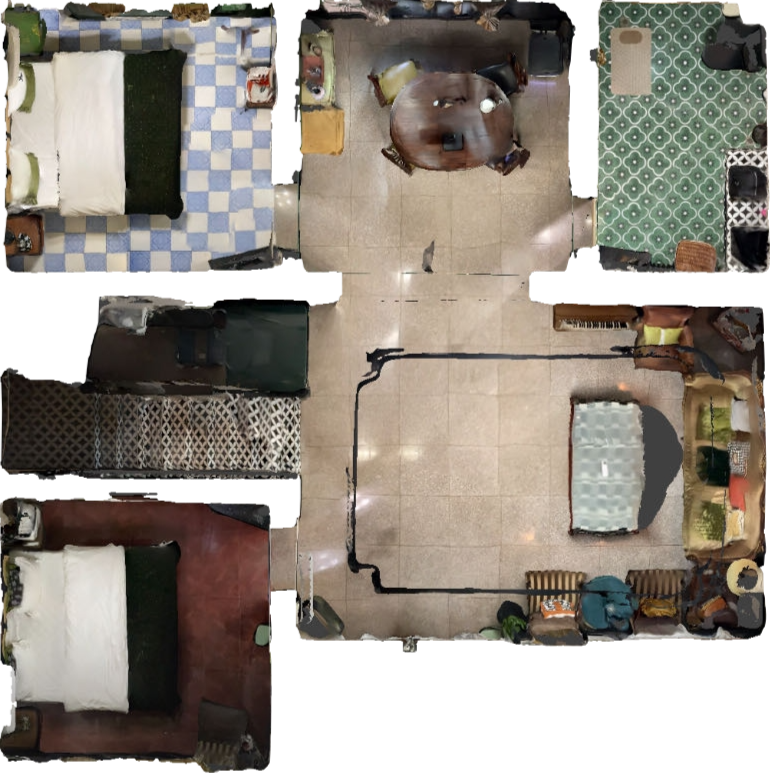}
        \caption{}
        \label{fig:real_scene}
    \end{subfigure}%
    \hspace{0.01\linewidth}
    \begin{subfigure}{0.36\linewidth}
        \includegraphics[width=\linewidth]{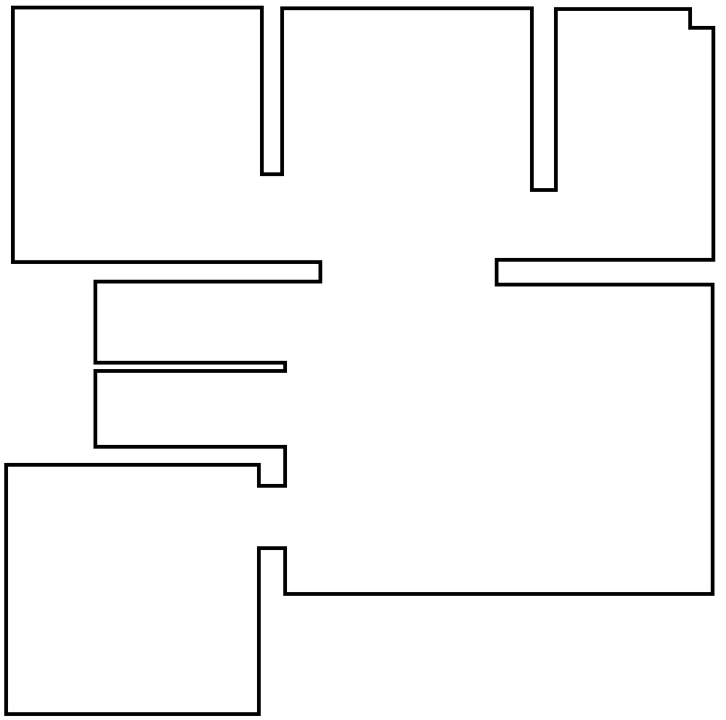}
        \caption{}
        \label{fig:real_floorplan}
    \end{subfigure}%
    \hspace{0.01\linewidth}
    \begin{subfigure}{0.23\linewidth}
        \includegraphics[width=\linewidth]{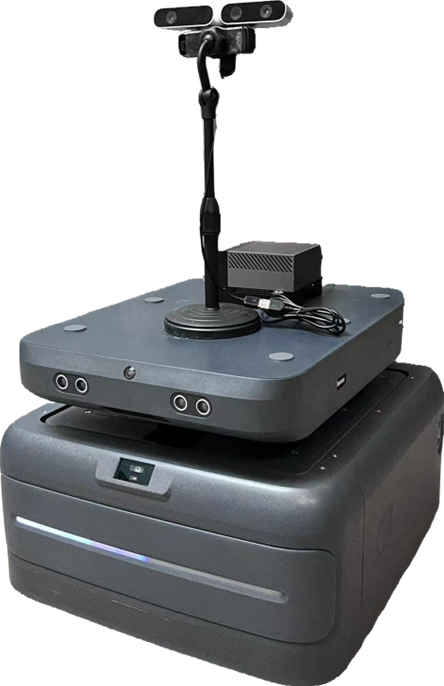}
        \caption{}
        \label{fig:real_robot}
    \end{subfigure}%
    \caption{\textbf{Real-world Deployment.} (a) The top view of the real scene and (b) the floor plan of the scene. (c) The robot platform features a wheeled base, an RGB camera, and an NVIDIA Jetson AGX Orin.}
    \label{fig:deployment}
    \vspace{-3pt}
\end{figure}

We conduct the experiment in an unseen apartment with an area of about $108 m^2$.  \cref{fig:real_scene} and \cref{fig:real_floorplan} show the scene layout and corresponding floor plan. \cref{fig:real_robot} shows the appearance of robot ``UP''. The agent stands $0.36$ meters tall, with a rectangular base measuring $0.46$ meters in length and $0.42$ meters in width. We additionally mounted an RGB camera on it and adjusted the camera height to $0.8m$. The apartment is pre-scanned, and we evaluate our Loc-\model (GT) in the scanned scene. The model generates actions based on RGB observations in simulation, which are then applied to ``UP'' and converted into linear and angular velocities using PD control in the real world. Our model achieves an inference rate of approximately $1.88$Hz when running on an NVIDIA Jetson AGX Orin.
To provide the current pose for the agent, we use a single-line LiDAR-based odometry algorithm. Before the testing, we manually annotate the agent's starting pose, represented as $\mathcal{T}_0$, in the scene coordinate system. During navigation, the LiDAR odometry continuously calculates the agent’s pose relative to $\mathcal{T}_0$ and converts it to the scene coordinate system. At time $t$, the agent pose $\mathcal{T}_t$ is calculated as $\mathcal{T}_t = \mathcal{T}_0 \circ \mathcal{T}_{t\rightarrow0}$, where $\mathcal{T}_{t\rightarrow0}$ represents the relative pose from $t$ to $0$ and $\circ$ represents the coordinate transformation operation.

\section{Conclusion}
In summary, this work introduces \task, the first to incorporate floor plans into embodied visual navigation. To tackle the \task, we develop \model, an efficient and effective diffusion policy framework integrating an explicit localization module, and curate a dataset for benchmarking. The superior performance of \model over comparison baselines highlights the promising potential of our approach. By presenting this practical setting and solution, we aim to inspire further research in the visual navigation community and contribute to advancements in the field.

\paragraph{Acknowledgement}
This project was supported by the National Natural Science Foundation of China (NSFC) under Grant No.62172043.
\\
\\
\\

\bibliography{reference}

\begin{thebibliography}{49}
\providecommand{\natexlab}[1]{#1}

\bibitem[{Anderson et~al.(2018{\natexlab{a}})Anderson, Chang, Chaplot, Dosovitskiy, Gupta, Koltun, Kosecka, Malik, Mottaghi, Savva et~al.}]{anderson2018evaluation}
Anderson, P.; Chang, A.; Chaplot, D.~S.; Dosovitskiy, A.; Gupta, S.; Koltun, V.; Kosecka, J.; Malik, J.; Mottaghi, R.; Savva, M.; et~al. 2018{\natexlab{a}}.
\newblock On evaluation of embodied navigation agents.
\newblock \emph{arXiv preprint arXiv:1807.06757}.

\bibitem[{Anderson et~al.(2018{\natexlab{b}})Anderson, Wu, Teney, Bruce, Johnson, S{\"u}nderhauf, Reid, Gould, and Van Den~Hengel}]{anderson2018vision}
Anderson, P.; Wu, Q.; Teney, D.; Bruce, J.; Johnson, M.; S{\"u}nderhauf, N.; Reid, I.; Gould, S.; and Van Den~Hengel, A. 2018{\natexlab{b}}.
\newblock Vision-and-language navigation: Interpreting visually-grounded navigation instructions in real environments.
\newblock In \emph{Conference on Computer Vision and Pattern Recognition (CVPR)}.

\bibitem[{Boniardi et~al.(2019{\natexlab{a}})Boniardi, Caselitz, K{\"u}mmerle, and Burgard}]{boniardi2019pose}
Boniardi, F.; Caselitz, T.; K{\"u}mmerle, R.; and Burgard, W. 2019{\natexlab{a}}.
\newblock A pose graph-based localization system for long-term navigation in CAD floor plans.
\newblock \emph{Robotics and Autonomous Systems}, 112: 84--97.

\bibitem[{Boniardi et~al.(2017)Boniardi, Caselitz, Kümmerle, and Burgard}]{boniardi2017robust}
Boniardi, F.; Caselitz, T.; Kümmerle, R.; and Burgard, W. 2017.
\newblock Robust LiDAR-based localization in architectural floor plans.
\newblock In \emph{International Conference on Intelligent Robots and Systems (IROS)}.

\bibitem[{Boniardi et~al.(2019{\natexlab{b}})Boniardi, Valada, Mohan, Caselitz, and Burgard}]{boniardi2019robot}
Boniardi, F.; Valada, A.; Mohan, R.; Caselitz, T.; and Burgard, W. 2019{\natexlab{b}}.
\newblock Robot Localization in Floor Plans Using a Room Layout Edge Extraction Network.
\newblock In \emph{International Conference on Intelligent Robots and Systems (IROS)}.

\bibitem[{Chaplot et~al.(2020{\natexlab{a}})Chaplot, Gandhi, Gupta, and Salakhutdinov}]{chaplot2020object}
Chaplot, D.~S.; Gandhi, D.; Gupta, A.; and Salakhutdinov, R. 2020{\natexlab{a}}.
\newblock Object Goal Navigation using Goal-Oriented Semantic Exploration.
\newblock In \emph{Advances in Neural Information Processing Systems (NeurIPS)}.

\bibitem[{Chaplot et~al.(2020{\natexlab{b}})Chaplot, Salakhutdinov, Gupta, and Gupta}]{chaplot2020neural}
Chaplot, D.~S.; Salakhutdinov, R.; Gupta, A.; and Gupta, S. 2020{\natexlab{b}}.
\newblock Neural topological slam for visual navigation.
\newblock In \emph{Conference on Computer Vision and Pattern Recognition (CVPR)}.

\bibitem[{Chen et~al.(2020)Chen, Jain, Schissler, Gari, Al-Halah, Ithapu, Robinson, and Grauman}]{chen2020soundspaces}
Chen, C.; Jain, U.; Schissler, C.; Gari, S. V.~A.; Al-Halah, Z.; Ithapu, V.~K.; Robinson, P.; and Grauman, K. 2020.
\newblock Soundspaces: Audio-visual navigation in 3d environments.
\newblock In \emph{European Conference on Computer Vision (ECCV)}.

\bibitem[{Chen et~al.(2021)Chen, Majumder, Ziad, Gao, Kumar~Ramakrishnan, and Grauman}]{chen2021learning}
Chen, C.; Majumder, S.; Ziad, A.-H.; Gao, R.; Kumar~Ramakrishnan, S.; and Grauman, K. 2021.
\newblock Learning to Set Waypoints for Audio-Visual Navigation.
\newblock In \emph{International Conference on Learning Representations (ICLR)}.

\bibitem[{Chen et~al.(2024)Chen, Wang, Vogel, and Pollefeys}]{chen2024f3loc}
Chen, C.; Wang, R.; Vogel, C.; and Pollefeys, M. 2024.
\newblock F$^3$Loc: Fusion and Filtering for Floorplan Localization.
\newblock In \emph{Conference on Computer Vision and Pattern Recognition (CVPR)}.

\bibitem[{Chi et~al.(2023)Chi, Feng, Du, Xu, Cousineau, Burchfiel, and Song}]{chi2023diffusion}
Chi, C.; Feng, S.; Du, Y.; Xu, Z.; Cousineau, E.; Burchfiel, B.; and Song, S. 2023.
\newblock Diffusion Policy: Visuomotor Policy Learning via Action Diffusion.
\newblock In \emph{Robotics: Science and Systems (RSS)}.

\bibitem[{Chu, Kim, and Chen(2015)}]{chu2015you}
Chu, H.; Kim, D.~K.; and Chen, T. 2015.
\newblock You are here: Mimicking the human thinking process in reading floor-plans.
\newblock In \emph{International Conference on Computer Vision (ICCV)}.

\bibitem[{Gadre et~al.(2023)Gadre, Wortsman, Ilharco, Schmidt, and Song}]{gadre2023cows}
Gadre, S.~Y.; Wortsman, M.; Ilharco, G.; Schmidt, L.; and Song, S. 2023.
\newblock Cows on pasture: Baselines and benchmarks for language-driven zero-shot object navigation.
\newblock In \emph{Conference on Computer Vision and Pattern Recognition (CVPR)}.

\bibitem[{Gan et~al.(2020)Gan, Zhang, Wu, Gong, and Tenenbaum}]{gan2020look}
Gan, C.; Zhang, Y.; Wu, J.; Gong, B.; and Tenenbaum, J.~B. 2020.
\newblock Look, listen, and act: Towards audio-visual embodied navigation.
\newblock In \emph{International Conference on Robotics and Automation (ICRA)}.

\bibitem[{Ho, Jain, and Abbeel(2020)}]{ho2020denoising}
Ho, J.; Jain, A.; and Abbeel, P. 2020.
\newblock Denoising Diffusion Probabilistic Models.
\newblock In \emph{Advances in Neural Information Processing Systems (NeurIPS)}.

\bibitem[{Howard-Jenkins and Prisacariu(2022)}]{howard2022lalaloc++}
Howard-Jenkins, H.; and Prisacariu, V.~A. 2022.
\newblock LaLaLoc++: Global floor plan comprehension for layout localisation in unvisited environments.
\newblock In \emph{European Conference on Computer Vision (ECCV)}.

\bibitem[{Howard-Jenkins, Ruiz-Sarmiento, and Prisacariu(2021)}]{howard2021lalaloc}
Howard-Jenkins, H.; Ruiz-Sarmiento, J.-R.; and Prisacariu, V.~A. 2021.
\newblock LaLaLoc: Latent Layout Localisation in Dynamic, Unvisited Environments.
\newblock In \emph{International Conference on Computer Vision (ICCV)}.

\bibitem[{Ito et~al.(2014)Ito, Endres, Kuderer, Tipaldi, Stachniss, and Burgard}]{ito2014w}
Ito, S.; Endres, F.; Kuderer, M.; Tipaldi, G.~D.; Stachniss, C.; and Burgard, W. 2014.
\newblock W-rgb-d: floor-plan-based indoor global localization using a depth camera and wifi.
\newblock In \emph{International Conference on Robotics and Automation (ICRA)}.

\bibitem[{Janner et~al.(2022)Janner, Du, Tenenbaum, and Levine}]{janner2022diffuser}
Janner, M.; Du, Y.; Tenenbaum, J.; and Levine, S. 2022.
\newblock Planning with Diffusion for Flexible Behavior Synthesis.
\newblock In \emph{International Conference on Machine Learning (ICML)}.

\bibitem[{Krantz et~al.(2023)Krantz, Gervet, Yadav, Wang, Paxton, Mottaghi, Batra, Malik, Lee, and Chaplot}]{krantz2023navigating}
Krantz, J.; Gervet, T.; Yadav, K.; Wang, A.; Paxton, C.; Mottaghi, R.; Batra, D.; Malik, J.; Lee, S.; and Chaplot, D.~S. 2023.
\newblock Navigating to objects specified by images.
\newblock In \emph{International Conference on Computer Vision (ICCV)}.

\bibitem[{Li et~al.(2021{\natexlab{a}})Li, Xia, Mart\'in-Mart\'in, Lingelbach, Srivastava, Shen, Vainio, Gokmen, Dharan, Jain, Kurenkov, Liu, Gweon, Wu, Fei-Fei, and Savarese}]{li2021igibson}
Li, C.; Xia, F.; Mart\'in-Mart\'in, R.; Lingelbach, M.; Srivastava, S.; Shen, B.; Vainio, K.~E.; Gokmen, C.; Dharan, G.; Jain, T.; Kurenkov, A.; Liu, K.; Gweon, H.; Wu, J.; Fei-Fei, L.; and Savarese, S. 2021{\natexlab{a}}.
\newblock iGibson 2.0: Object-centric simulation for robot learning of everyday household tasks.
\newblock In \emph{Conference on Robot Learning (CoRL)}.

\bibitem[{Li et~al.(2021{\natexlab{b}})Li, Chan, Le~Chan, Li, Wan, and Yau}]{li2021cognitive}
Li, J.; Chan, C.~L.; Le~Chan, J.; Li, Z.; Wan, K.~W.; and Yau, W.~Y. 2021{\natexlab{b}}.
\newblock Cognitive navigation for indoor environment using floorplan.
\newblock In \emph{International Conference on Intelligent Robots and Systems (IROS)}.

\bibitem[{Li, Ang, and Rus(2020)}]{li2020online}
Li, Z.; Ang, M.~H.; and Rus, D. 2020.
\newblock Online localization with imprecise floor space maps using stochastic gradient descent.
\newblock In \emph{International Conference on Intelligent Robots and Systems (IROS)}.

\bibitem[{Loshchilov, Hutter et~al.(2017)}]{loshchilov2017fixing}
Loshchilov, I.; Hutter, F.; et~al. 2017.
\newblock Fixing weight decay regularization in adam.
\newblock \emph{arXiv preprint arXiv:1711.05101}, 5.

\bibitem[{Mendez et~al.(2020)Mendez, Hadfield, Pugeault, and Bowden}]{mendez2020sedar}
Mendez, O.; Hadfield, S.; Pugeault, N.; and Bowden, R. 2020.
\newblock SeDAR: reading floorplans like a human—using deep learning to enable human-inspired localisation.
\newblock \emph{International Journal of Computer Vision (IJCV)}, 128(5): 1286--1310.

\bibitem[{Min et~al.(2021)Min, Chaplot, Ravikumar, Bisk, and Salakhutdinov}]{min2021film}
Min, S.~Y.; Chaplot, D.~S.; Ravikumar, P.; Bisk, Y.; and Salakhutdinov, R. 2021.
\newblock Film: Following instructions in language with modular methods.
\newblock \emph{arXiv preprint arXiv:2110.07342}.

\bibitem[{Min et~al.(2022)Min, Khosravan, Bessinger, Narayana, Kang, Dunn, and Boyadzhiev}]{min2022laser}
Min, Z.; Khosravan, N.; Bessinger, Z.; Narayana, M.; Kang, S.~B.; Dunn, E.; and Boyadzhiev, I. 2022.
\newblock Laser: Latent space rendering for 2d visual localization.
\newblock In \emph{Conference on Computer Vision and Pattern Recognition (CVPR)}.

\bibitem[{Nichol and Dhariwal(2021)}]{nichol2021improved}
Nichol, A.~Q.; and Dhariwal, P. 2021.
\newblock Improved denoising diffusion probabilistic models.
\newblock In \emph{International Conference on Machine Learning (ICML)}.

\bibitem[{Partsey et~al.(2022)Partsey, Wijmans, Yokoyama, Dobosevych, Batra, and Maksymets}]{partsey2022mapping}
Partsey, R.; Wijmans, E.; Yokoyama, N.; Dobosevych, O.; Batra, D.; and Maksymets, O. 2022.
\newblock Is mapping necessary for realistic pointgoal navigation?
\newblock In \emph{Conference on Computer Vision and Pattern Recognition (CVPR)}.

\bibitem[{Ramakrishnan et~al.(2022)Ramakrishnan, Chaplot, Al-Halah, Malik, and Grauman}]{ramakrishnan2022poni}
Ramakrishnan, S.~K.; Chaplot, D.~S.; Al-Halah, Z.; Malik, J.; and Grauman, K. 2022.
\newblock PONI: Potential Functions for ObjectGoal Navigation With Interaction-Free Learning.
\newblock In \emph{Conference on Computer Vision and Pattern Recognition (CVPR)}.

\bibitem[{Ramakrishnan et~al.(2021)Ramakrishnan, Gokaslan, Wijmans, Maksymets, Clegg, Turner, Undersander, Galuba, Westbury, Chang et~al.}]{ramakrishnan2021habitat}
Ramakrishnan, S.~K.; Gokaslan, A.; Wijmans, E.; Maksymets, O.; Clegg, A.; Turner, J.; Undersander, E.; Galuba, W.; Westbury, A.; Chang, A.~X.; et~al. 2021.
\newblock Habitat-matterport 3d dataset (hm3d): 1000 large-scale 3d environments for embodied ai.
\newblock \emph{arXiv preprint arXiv:2109.08238}.

\bibitem[{Savva et~al.(2019)Savva, Kadian, Maksymets, Zhao, Wijmans, Jain, Straub, Liu, Koltun, Malik, Parikh, and Batra}]{savva2019habitat}
Savva, M.; Kadian, A.; Maksymets, O.; Zhao, Y.; Wijmans, E.; Jain, B.; Straub, J.; Liu, J.; Koltun, V.; Malik, J.; Parikh, D.; and Batra, D. 2019.
\newblock Habitat: A platform for embodied ai research.
\newblock In \emph{International Conference on Computer Vision (ICCV)}.

\bibitem[{Setalaphruk et~al.(2003)Setalaphruk, Ueno, Kume, Kono, and Kidode}]{setalaphruk2003robot}
Setalaphruk, V.; Ueno, A.; Kume, I.; Kono, Y.; and Kidode, M. 2003.
\newblock Robot navigation in corridor environments using a sketch floor map.
\newblock In \emph{International Symposium on Computational Intelligence in Robotics and Automation.}

\bibitem[{Shah et~al.(2021)Shah, Eysenbach, Kahn, Rhinehart, and Levine}]{shah2021ving}
Shah, D.; Eysenbach, B.; Kahn, G.; Rhinehart, N.; and Levine, S. 2021.
\newblock ViNG: Learning Open-World Navigation with Visual Goals.
\newblock In \emph{International Conference on Robotics and Automation (ICRA)}.

\bibitem[{Shah et~al.(2023)Shah, Sridhar, Dashora, Stachowicz, Black, Hirose, and Levine}]{shah2023vint}
Shah, D.; Sridhar, A.; Dashora, N.; Stachowicz, K.; Black, K.; Hirose, N.; and Levine, S. 2023.
\newblock Vi{NT}: A Foundation Model for Visual Navigation.
\newblock In \emph{Conference on Robot Learning (CoRL)}.

\bibitem[{Sohl-Dickstein et~al.(2015)Sohl-Dickstein, Weiss, Maheswaranathan, and Ganguli}]{sohl2015deep}
Sohl-Dickstein, J.; Weiss, E.; Maheswaranathan, N.; and Ganguli, S. 2015.
\newblock Deep unsupervised learning using nonequilibrium thermodynamics.
\newblock In \emph{International Conference on Machine Learning (ICML)}.

\bibitem[{Song et~al.(2023)Song, Wu, Washington, Sadler, Chao, and Su}]{song2023llm}
Song, C.~H.; Wu, J.; Washington, C.; Sadler, B.~M.; Chao, W.-L.; and Su, Y. 2023.
\newblock Llm-planner: Few-shot grounded planning for embodied agents with large language models.
\newblock In \emph{International Conference on Computer Vision (ICCV)}.

\bibitem[{Sridhar et~al.(2023)Sridhar, Shah, Glossop, and Levine}]{sridhar2023nomad}
Sridhar, A.; Shah, D.; Glossop, C.; and Levine, S. 2023.
\newblock Nomad: Goal masked diffusion policies for navigation and exploration.
\newblock \emph{arXiv preprint arXiv:2310.07896}.

\bibitem[{Tan and Le(2019)}]{tan2019efficientnet}
Tan, M.; and Le, Q. 2019.
\newblock Efficientnet: Rethinking model scaling for convolutional neural networks.
\newblock In \emph{International Conference on Machine Learning (ICML)}.

\bibitem[{Tang et~al.(2022)Tang, Du, Yu, and Yang}]{tang2022monocular}
Tang, T.; Du, H.; Yu, X.; and Yang, Y. 2022.
\newblock Monocular camera-based point-goal navigation by learning depth channel and cross-modality pyramid fusion.
\newblock In \emph{AAAI Conference on Artificial Intelligence (AAAI)}.

\bibitem[{Vaswani et~al.(2017)Vaswani, Shazeer, Parmar, Uszkoreit, Jones, Gomez, Kaiser, and Polosukhin}]{vaswani2017attention}
Vaswani, A.; Shazeer, N.; Parmar, N.; Uszkoreit, J.; Jones, L.; Gomez, A.~N.; Kaiser, {\L}.; and Polosukhin, I. 2017.
\newblock Attention is all you need.
\newblock In \emph{Advances in Neural Information Processing Systems (NeurIPS)}.

\bibitem[{Wang et~al.(2023{\natexlab{a}})Wang, Liang, Van~Gool, and Wang}]{wang2023dreamwalker}
Wang, H.; Liang, W.; Van~Gool, L.; and Wang, W. 2023{\natexlab{a}}.
\newblock Dreamwalker: Mental planning for continuous vision-language navigation.
\newblock In \emph{International Conference on Computer Vision (ICCV)}.

\bibitem[{Wang et~al.(2023{\natexlab{b}})Wang, Wang, Liang, Hoi, Shen, and Gool}]{wang2023active}
Wang, H.; Wang, W.; Liang, W.; Hoi, S.~C.; Shen, J.; and Gool, L.~V. 2023{\natexlab{b}}.
\newblock Active perception for visual-language navigation.
\newblock \emph{International Journal of Computer Vision (IJCV)}, 131(3): 607--625.

\bibitem[{Wang, Marcotte, and Olson(2019)}]{wang2019glfp}
Wang, X.; Marcotte, R.~J.; and Olson, E. 2019.
\newblock GLFP: Global localization from a floor plan.
\newblock In \emph{International Conference on Intelligent Robots and Systems (IROS)}.

\bibitem[{Wijmans et~al.(2019)Wijmans, Kadian, Morcos, Lee, Essa, Parikh, Savva, and Batra}]{wijmans2019dd}
Wijmans, E.; Kadian, A.; Morcos, A.; Lee, S.; Essa, I.; Parikh, D.; Savva, M.; and Batra, D. 2019.
\newblock Dd-ppo: Learning near-perfect pointgoal navigators from 2.5 billion frames.
\newblock \emph{arXiv preprint arXiv:1911.00357}.

\bibitem[{Xia et~al.(2018)Xia, Zamir, He, Sax, Malik, and Savarese}]{xia2018gibson}
Xia, F.; Zamir, A.~R.; He, Z.; Sax, A.; Malik, J.; and Savarese, S. 2018.
\newblock Gibson env: Real-world perception for embodied agents.
\newblock In \emph{Conference on Computer Vision and Pattern Recognition (CVPR)}.

\bibitem[{Yadav et~al.(2023)Yadav, Majumdar, Ramrakhya, Yokoyama, Baevski, Kira, Maksymets, and Batra}]{yadav2023ovrl}
Yadav, K.; Majumdar, A.; Ramrakhya, R.; Yokoyama, N.; Baevski, A.; Kira, Z.; Maksymets, O.; and Batra, D. 2023.
\newblock Ovrl-v2: A simple state-of-art baseline for imagenav and objectnav.
\newblock \emph{arXiv preprint arXiv:2303.07798}.

\bibitem[{Zhao et~al.(2021)Zhao, Agrawal, Batra, and Schwing}]{zhao2021surprising}
Zhao, X.; Agrawal, H.; Batra, D.; and Schwing, A.~G. 2021.
\newblock The surprising effectiveness of visual odometry techniques for embodied pointgoal navigation.
\newblock In \emph{International Conference on Computer Vision (ICCV)}.

\bibitem[{Zhu et~al.(2017)Zhu, Mottaghi, Kolve, Lim, Gupta, Fei-Fei, and Farhadi}]{zhu2017target}
Zhu, Y.; Mottaghi, R.; Kolve, E.; Lim, J.~J.; Gupta, A.; Fei-Fei, L.; and Farhadi, A. 2017.
\newblock Target-driven visual navigation in indoor scenes using deep reinforcement learning.
\newblock In \emph{International Conference on Robotics and Automation (ICRA)}.

\end{thebibliography}

\clearpage
\appendix
\renewcommand\thefigure{A\arabic{figure}}
\setcounter{figure}{0}
\renewcommand\thetable{A\arabic{table}}
\setcounter{table}{0}
\renewcommand\theequation{A\arabic{equation}}
\setcounter{equation}{0}
\pagenumbering{arabic}
\renewcommand*{\thepage}{A\arabic{page}}

\begin{strip}%
\centering
\textbf{\LARGE FloNa: Floor Plan Guided Embodied Visual Navigation\\
\textit{(Supplementary Material)}}
\end{strip}

\section{Distinctions from PointNav}
The distinctions between FloNa and PointNav \cite{anderson2018evaluation} lie mainly in two aspects. First, PointNav typically involves time-consuming exploration of the environment, either explicitly or implicitly, due to the lack of global information. In contrast, FloNa leverages the geometric information embedded in floor plans to enable more efficient navigation. Secondly, PointNav defines the goal in relative coordinates, whereas FloNa specifies the goal in the global coordinate system without including information about the relative position between the agent and the goal, requiring the model to incorporate the localization capability.

\section{Dataset Details}
\begin{figure*}[t!]
  \centering
  \captionsetup[subfloat]{labelformat=empty}
  \subfloat[]{\parbox[b]{0.23\textwidth}{\centering Scene \\ \vspace{0.5em}\includegraphics[width=\linewidth]{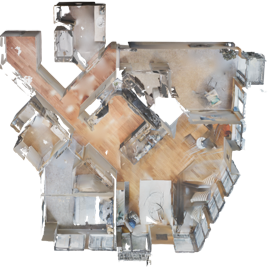}}}\hfill
  \subfloat[]{\parbox[b]{0.23\textwidth}{\centering Floorplan \\ \vspace{0.5em}\includegraphics[width=\linewidth]{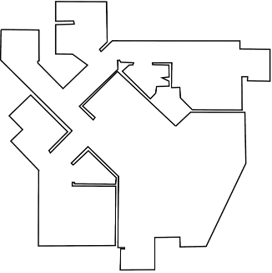}}}\hfill
  \subfloat[]{\parbox[b]{0.23\textwidth}{\centering Rough \\ \vspace{0.5em}\includegraphics[width=\linewidth]{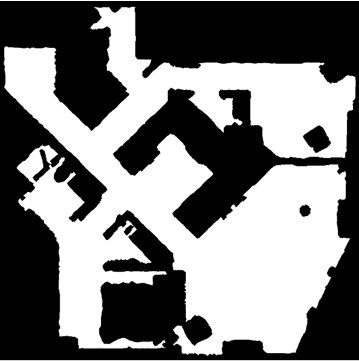}}}\hfill
  \subfloat[]{\parbox[b]{0.23\textwidth}{\centering Fine \\ \vspace{0.5em}\includegraphics[width=\linewidth]{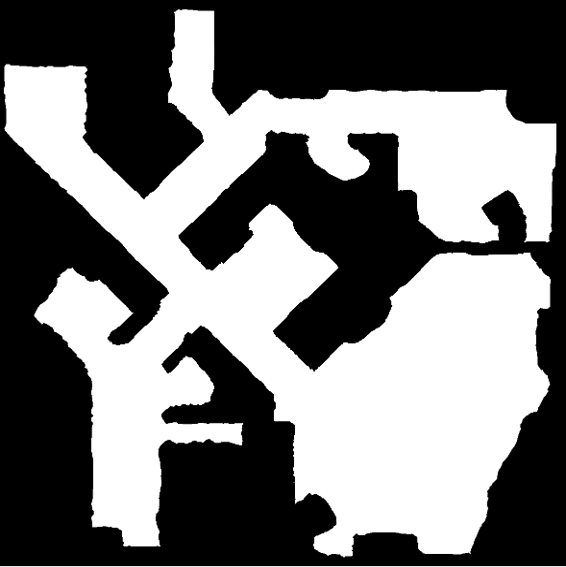}}}\\
  \subfloat[]{\includegraphics[width=0.23\textwidth]{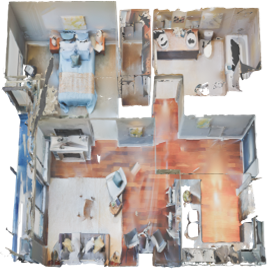}}\hfill
  \subfloat[]{\includegraphics[width=0.23\textwidth]{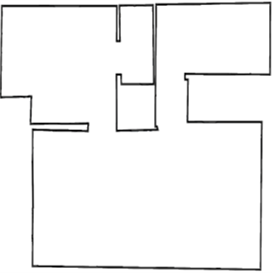}}\hfill
  \subfloat[]{\includegraphics[width=0.23\textwidth]{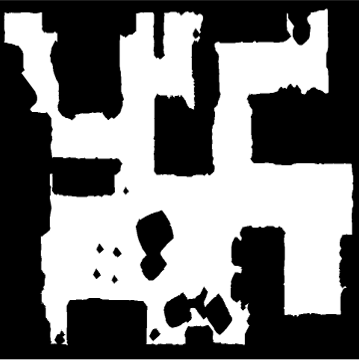}}\hfill
  \subfloat[]{\includegraphics[width=0.23\textwidth]{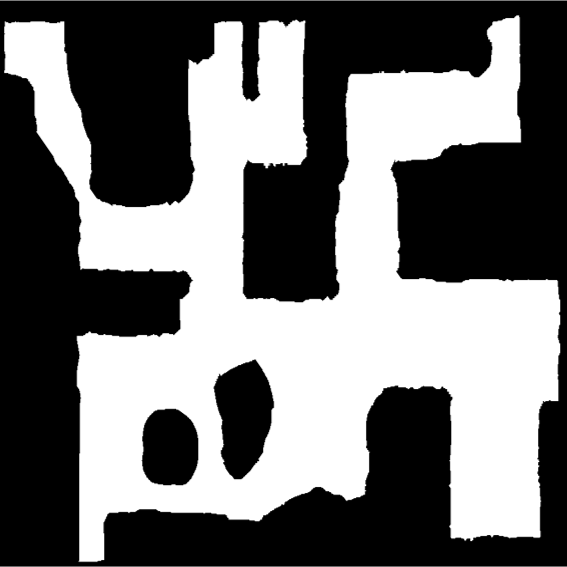}}\\
  \subfloat[]{\includegraphics[width=0.23\textwidth]{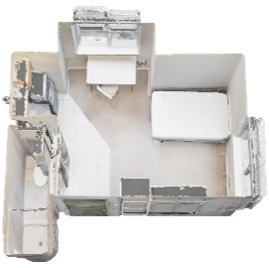}}\hfill
  \subfloat[]{\includegraphics[width=0.23\textwidth]{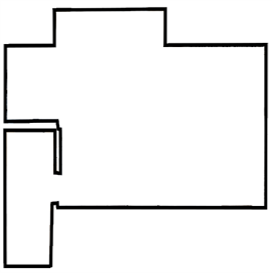}}\hfill
  \subfloat[]{\includegraphics[width=0.23\textwidth]{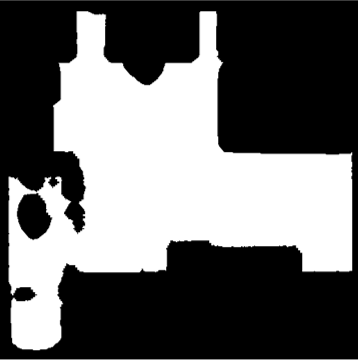}}\hfill
  \subfloat[]{\includegraphics[width=0.23\textwidth]{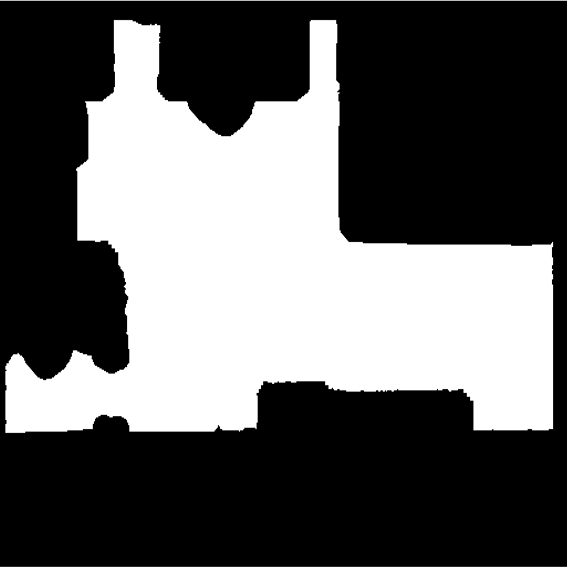}}
  \caption{Three scenes and their corresponding floor plans and navigable maps. From top to bottom, the scenes represent a large scene, a medium scene, and a small scene. The first and second columns display the 3D models of the scenes and their corresponding floor plans, respectively. The third column shows the coarse navigable maps generated in the first stage, while the fourth column presents the refined navigable maps manually adjusted in the second stage.}
  \label{fig:dataset}
\end{figure*}

We collect the dataset in 117 static indoor scenes, which are split into 67 for training and 50 for testing. Each scene is accompanied by a floor plan, a navigable map, and multiple episodes of data. Depending on the scene size, we collected 150 episodes in small scenes (with areas less than 20 $m^2$), 180 in medium scenes (areas between 20 and 80 $m^2$), and 200 in large scenes (areas greater than 80 $m^2$). Each episode includes a trajectory and the observations captured throughout. In total, the dataset comprises 20,214 navigation episodes, resulting in 3,312,480 RGB images. \cref{fig:dataset} shows three examples of our scenes as well as corresponding floor plans and navigable maps.  

\paragraph{Navigable Maps}
Scene and corresponding floor plans are provided by Gibson \cite{xia2018gibson} and \citet{chen2024f3loc}, respectively. For the navigable maps, we employ a two-stage generation process. In the first stage, we generate a rough navigable map by selecting meshes with heights between 0.15 $m$ and 1.70 $m$.  These meshes are treated as obstacles and marked in black, while the remaining non-obstacle areas are marked in white. However, due to imperfect reconstruction (\eg, incomplete chair leg reconstruction or incorrect floor reconstruction), some unnavigable areas may be mistakenly marked as navigable, and vice versa. Therefore, in the second stage, we manually refine the rough map to get a fine navigable map. 

\paragraph{Episodes Sampling}
For each episode, we first dilate the unnavigable areas to approximate the radius of the robots, ensuring collision-free travel. Using this dilated map, we randomly select two points at least 3 meters apart within the navigable area as the starting and target points. We then use the A* algorithm to generate the shortest path, represented as a sequence of pixel coordinates. These pixel coordinates are then converted into coordinates in the scene coordinate system (in meters) to form the trajectory data. To ensure the smoothness of the trajectory, the orientation at each position is determined by the location of the sixth future point, \ie, $r_i = p_{i+6} - p_{i}$, where $r_i$ and $p_i$ denote the orientation and position of each point $i$. We track this trajectory in the iGibson simulation environment and render the corresponding RGB image with a resolution of $512\times512$ at each position. Notably, the camera's height, pitch, and roll are kept constant (height at 0.05 $m$, and roll and pitch at 0$^\circ$) during the travel in the simulation environment.

\begin{table}[t!]
    \centering
    \small
    \caption{\textbf{SR($\%$) and SPL($\%$) of Loc-\model under different F$^3$ models}. The \textbf{bold} number indicates the best result. F$^{3*}$ refers to F$^3$Loc trained from scratch with data collected at 80cm height, while F$^3$ utilizes the pre-trained model.}
    \vspace{-3pt}
    \label{tab:f3results}
    \resizebox{\linewidth}{!}{%
    \begin{tabular}{ccccccccc}%
        \toprule
        \multicolumn{3}{c}{\multirow{3}[3]{*}{Method}}& \multicolumn{3}{c}{Loc-\model (F$^3$)}     & \multicolumn{3}{c}{Loc-\model (F$^{3*}$)} \\
        \cmidrule{4-9}
        &&    & \multicolumn{3}{c}{$\tau_d (m)$}     & \multicolumn{3}{c}{$\tau_d (m)$} \\
        \cmidrule(rr){4-6}\cmidrule(l){7-9}
        &&    & 0.25 & 0.30 & 0.35                   & 0.25 & 0.30 & 0.35\\
        \midrule
        \multicolumn{1}{c|}{\multirow{4}{*}{SR(\%) $\uparrow$}} & \multicolumn{1}{c}{\multirow{4}{*}{$\tau_c$}}   & 10       & 4.80 & 6.00 & 6.40    & \textbf{6.60} & \textbf{7.60} & \textbf{8.80}   \\
        \multicolumn{1}{c|}{} & \multicolumn{1}{c}{}     & 30       & 11.6 & 14.0 & 15.0    & \textbf{12.6} & \textbf{15.4} & \textbf{17.0} \\
        \multicolumn{1}{c|}{} & \multicolumn{1}{c}{}     & 50       & 17.8 & 20.6 & 22.4    & \textbf{18.6} & \textbf{21.8} & \textbf{23.8} \\
        \multicolumn{1}{c|}{} & \multicolumn{1}{c}{}     & $\infty$ & \textbf{26.8} & 30.8 & 33.0    & 26.6    & \textbf{32.2} & \textbf{33.8} \\
        \midrule
        \multicolumn{1}{c|}{\multirow{4}{*}{SPL(\%) $\uparrow$}} & \multicolumn{1}{c}{\multirow{4}{*}{$\tau_c$}}   & 10       & 4.02 & 5.24 & 5.54    & \textbf{5.50} & \textbf{6.50} & \textbf{7.20}   \\
        \multicolumn{1}{c|}{} & \multicolumn{1}{c}{}     & 30       & 6.40 & 8.25 & 8.98    & \textbf{7.80} & \textbf{9.40} & \textbf{10.3} \\
        \multicolumn{1}{c|}{} & \multicolumn{1}{c}{}     & 50       & 8.07 & 9.98 & 10.8    & \textbf{9.10} & \textbf{10.8} & \textbf{11.8} \\
        \multicolumn{1}{c|}{} & \multicolumn{1}{c}{}     & $\infty$ & 9.58 & 11.7 & 12.5    & \textbf{10.4}    & \textbf{12.7} & \textbf{13.4} \\
        \bottomrule
    \end{tabular}
    }
\end{table}

\begin{table}[t!]
    \centering
    \small
    \caption{\textbf{SR($\%$) and SPL($\%$) of Loc-Diff(GT) and Loc-\model (GT)}. The \textbf{bold} number indicates the best result. }
    \vspace{-3pt}
    \label{tab:w/o floor plan}
    \resizebox{\linewidth}{!}{%
    \begin{tabular}{ccccccccc}%
        \toprule
        \multicolumn{3}{c}{\multirow{3}[3]{*}{Method}}& \multicolumn{3}{c}{Loc-Diff (GT)}     & \multicolumn{3}{c}{Loc-\model (GT)} \\
        \cmidrule{4-9}
        &&    & \multicolumn{3}{c}{$\tau_d (m)$}     & \multicolumn{3}{c}{$\tau_d (m)$} \\
        \cmidrule(rr){4-6}\cmidrule(l){7-9}
        &&    & 0.25 & 0.30 & 0.35                   & 0.25 & 0.30 & 0.35\\
        \midrule
        \multicolumn{1}{c|}{\multirow{4}{*}{SR(\%) $\uparrow$}} & \multicolumn{1}{c}{\multirow{4}{*}{$\tau_c$}}   & 10       & 35.0 & 38.0 & 38.2    & \textbf{39.0} & \textbf{44.2} & \textbf{44.6}   \\
        \multicolumn{1}{c|}{} & \multicolumn{1}{c}{}     & 30       & 51.6 & 56.8 & 57.0    & \textbf{53.4} & \textbf{60.8} & \textbf{61.0} \\
        \multicolumn{1}{c|}{} & \multicolumn{1}{c}{}     & 50       & 57.0 & 63.0 & 49.6    & \textbf{59.2} & \textbf{68.4} & \textbf{68.4} \\
        \multicolumn{1}{c|}{} & \multicolumn{1}{c}{}     & $\infty$ & 44.2 & 51.1 & 52.3    & \textbf{66.0}    & \textbf{75.8} & \textbf{76.0} \\
        \midrule
        \multicolumn{1}{c|}{\multirow{4}{*}{SPL(\%) $\uparrow$}} & \multicolumn{1}{c}{\multirow{4}{*}{$\tau_c$}}   & 10       & 32.1 & 35.0 & 35.7    & \textbf{36.3} & \textbf{41.4} & \textbf{42.1}   \\
        \multicolumn{1}{c|}{} & \multicolumn{1}{c}{}     & 30       & 40.6 & 44.9 & 45.8    & \textbf{44.1} & \textbf{50.4} & \textbf{51.2} \\
        \multicolumn{1}{c|}{} & \multicolumn{1}{c}{}     & 50       & 42.5 & 47.1 & 48.0    & \textbf{46.6} & \textbf{53.4} & \textbf{54.1} \\
        \multicolumn{1}{c|}{} & \multicolumn{1}{c}{}     & $\infty$ & 45.0 & 49.8    & 50.6    & \textbf{48.2}    & \textbf{55.2} & \textbf{56.0} \\
        \bottomrule
    \end{tabular}
    }
\end{table}

\section{Additional Experimental Results}

\paragraph{Re-train F$^3$Loc for Loc-\model (F$^3$)}
In our main experiments, the F$^3$Loc is pre-trained on data collected at the height of 175cm, while our images are collected at the height of 80cm. This discrepancy significantly impacts localization. Thus, we re-train F$^3$Loc using approximately 40k image-position pairs collected at the height of 80cm, resulting in a higher success rate, as shown in \cref{tab:f3results}. The definitions of $\tau_c$ and $\tau_d$ can be found in main experiments. The results suggest that improved localization accuracy leads to better navigation performance.

\paragraph{Loc-FloDiff (GT) without Floor Plan}
To validate the impact of the floor plan, we train an additional model, Loc-Diff, which shares the same architecture with Loc-\model. The only difference is that the floor plan embedding was masked in Loc-Diff during training. Consistently, we train this model for 5 epochs. The results are shown in \cref{tab:w/o floor plan}. The higher SR and SPL achieved by Loc-\model (GT) highlight the effectiveness of the floor plan in facilitating more efficient navigation.

\begin{figure}[t!]
    \begin{subfigure}{0.49\linewidth}
        \includegraphics[width=\linewidth]{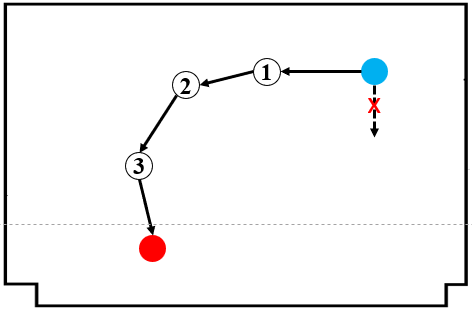}
        \caption{}
        \label{fig:floorplan}
    \end{subfigure}%
    \hspace{0.01\linewidth}
    \begin{subfigure}{0.49\linewidth}
        \includegraphics[width=\linewidth,height=0.67\linewidth]{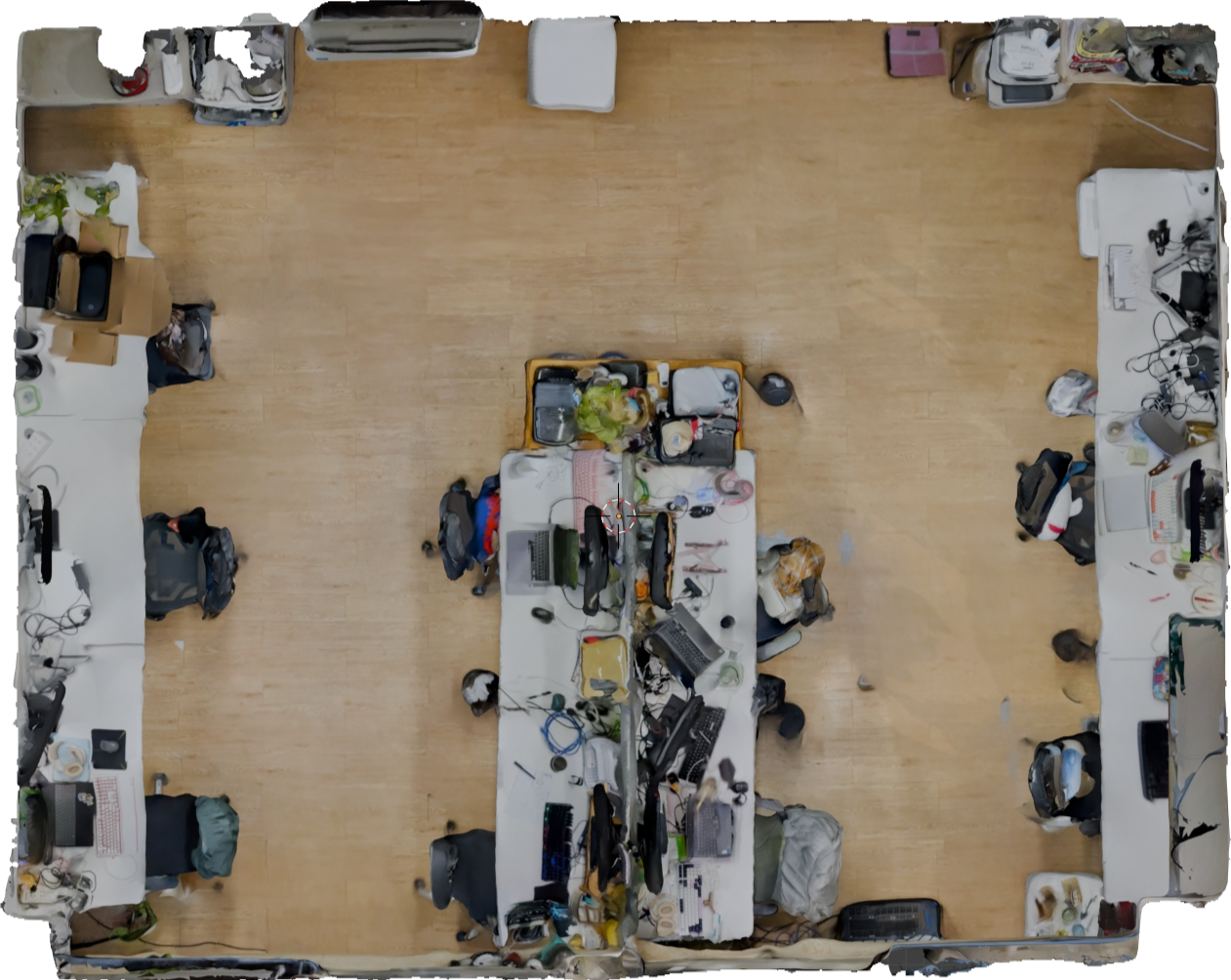}
        \caption{}
        \label{fig:bev}
    \end{subfigure} \\
    \begin{subfigure}{0.18\linewidth}
        \includegraphics[width=\linewidth]{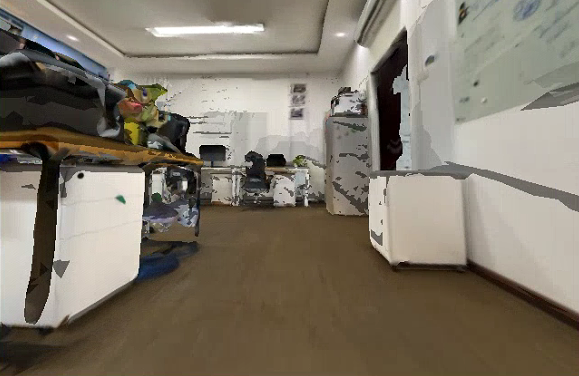}
        \caption*{start point}
    \end{subfigure}%
    \hspace{0.01\linewidth}
    \begin{subfigure}{0.18\linewidth}
        \includegraphics[width=\linewidth]{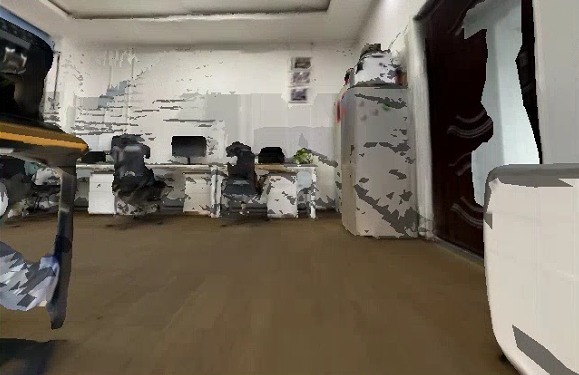}
        \caption*{1}        
    \end{subfigure}
    \hspace{0.01\linewidth}
    \begin{subfigure}{0.18\linewidth}
        \includegraphics[width=\linewidth]{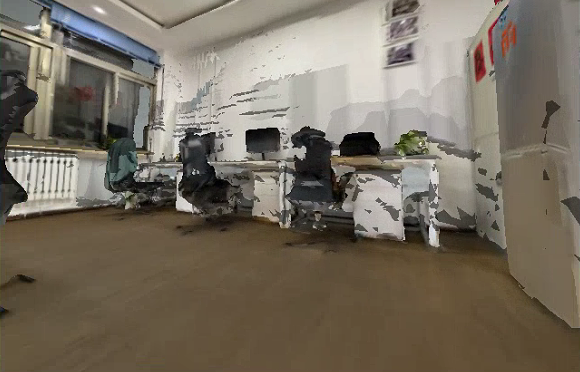}
        \caption*{2}        
    \end{subfigure}
    \hspace{0.01\linewidth}
    \begin{subfigure}{0.18\linewidth}
        \includegraphics[width=\linewidth]{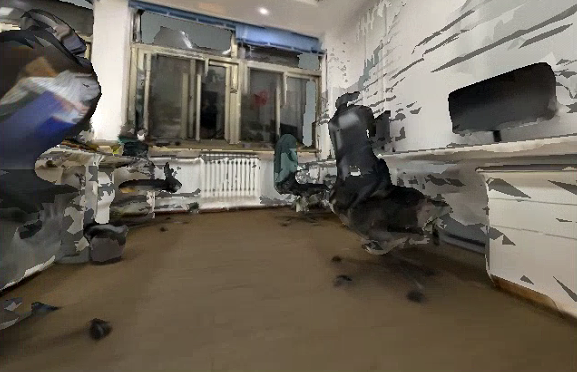}
        \caption*{3}        
    \end{subfigure}
    \hspace{0.01\linewidth}
    \begin{subfigure}{0.18\linewidth}
        \includegraphics[width=\linewidth]{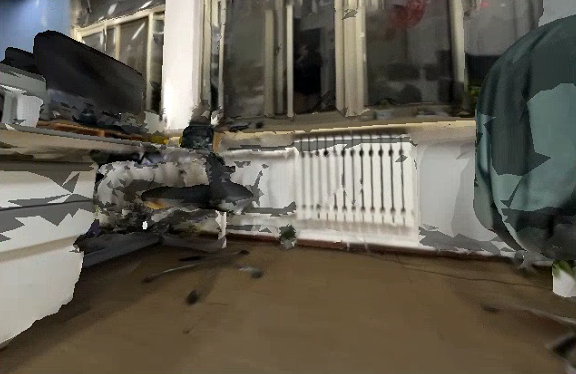}
        \caption*{goal}        
    \end{subfigure}
    \caption{\textbf{Generalizablity of Loc-\model (GT).} (a) The floor plan of an unseen office environment and an example of one episode. The red point indicates the goal, and the blue point indicates the starting position. (b) The top-down view of the office. The bottom row shows the observation sequence at the corresponding point.}
    \label{fig:quality}
\end{figure}

\paragraph{Qualitative Experiments}
To evaluate the generalization capability, we futher scan a crowded indoor office environment and test our model in it. The floor plan and the top-down view of the environment are shown in \cref{fig:quality} (a) and (b), respectively.  In this unseen environment, our model is still capable of performing navigation tasks effectively. \cref{fig:quality} (a) illustrates one of the test episodes, where the red point indicates the goal, and the blue point indicates the starting position. Our model can ``intelligently'' navigate around tables and chairs, even though these obstacles are not depicted in the floor plan. Additionally, at the starting position, the agent does not turn to the left, demonstrating that the model can leverage the global information provided by the floor plan to make correct decisions.

\section{Limitations and Future Works}

In this initial exploration, we fuse the floor plan with the image modality. In future work, our aim is to (i) enable the model to understand various forms of floor plans, \eg,  hand-drawn sketches, (ii) integrate additional modalities, such as language, sound, and objects, with floor plans to further enhance the functionality of floor plan-based navigation tasks, (iii) achieve effective navigation even with coarse localization inputs, and (iv) further improve obstacle avoidance capability of the policy, increasing success rates in challenging episodes. 

\end{document}